\documentclass[]{interact}
\usepackage{hyperref} 
\usepackage{epstopdf}
\usepackage[caption=false]{subfig}
\usepackage[linesnumbered, ruled]{algorithm2e}
\usepackage[numbers,sort&compress]{natbib}
\usepackage{xcolor}
\usepackage{multirow}
\usepackage{pdflscape}
\usepackage{multirow}
\usepackage{graphicx}

\usepackage{booktabs}
\usepackage[pagewise,mathlines]{lineno}

\bibpunct[, ]{[}{]}{,}{n}{,}{,}
\renewcommand\bibfont{\fontsize{10}{12}\selectfont}

\theoremstyle{plain}

\theoremstyle{definition}

\theoremstyle{remark}

\newif\ifblind
 \blindfalse

\begin{document}


\title{Knockoffs-based False Discovery Rate Control and Simplification for Deep Neural Networks}

\ifblind
\author{
\name{Authors}
\affil{Affiliation omitted for blind review}
}
\else
\author{
\name{Wenyu Liao, Yiqing Shi, and Fang Xie\thanks{CONTACT Fang Xie. Email: fangxie@bnbu.edu.cn}}
\affil{Guangdong Provincial Key Laboratory of IRADS, Beijing Normal-Hong Kong Baptist University,
Zhuhai 519087, China}
}
\fi
\maketitle

\begin{abstract}
The deep neural network is a widely used framework in machine learning that has been widely applied in various fields. However, deep neural networks often involve a large number of parameters and inputs, many of which may be irrelevant to the goal or true output. These parameters and \textcolor{black}{input variables} not only increase computational complexity, but also contribute to additional computational cost. One solution to this problem is knockoff methods, which have proven successful in controlling false discovery rates in high-dimensional regression. Building on the knockoff methods and using the regularised neural network, this paper proposes three variable screening methods under the condition of controlling false discovery rates: \textit{one layer filter}, \textit{multiple layers filter}, and \textit{variable weight aggregation filter}. In comparison with existing algorithms, we find that our algorithms show satisfactory performance.
\end{abstract}


\section{Introduction}\label{sec:Introduction}
In modern environments, many variables are typically collected, but it is difficult to know which ones are important to the response. One of our main objectives in this article is to address the variable screening problem.

By controlling the false discovery rate (FDR), it is possible to screen the variables well. Currently, there are many methods to control the error rate, such as knockoffs \cite{barber2015}, but most of these methods can only be used in linear models. For nonlinear models, the methods proposed in \cite{chen2021}, \cite{Dinh2020} and \cite{Zhang2019} can effectively select variables and deal with nonlinear issues. However, they do not control FDR and have high model training complexity. \cite{Yasuda2022} introduced a feature selection algorithm called sequential attention, which achieves remarkable results in neural networks. However, it relies on the attention mechanism, which may overlook the marginal contributions of features. This dependency may lead to the selection of redundant features, or the omission of features that would be informative when considered independently. \cite{kruz2022} is crucial for variable selection in high-dimensional settings, although the implementation and parameterisation of the model is somewhat complicated.

As we know, neural networks have better performance when dealing with non-linear complex classification problems. But for neural networks, there are few methods for variable selection under FDR.  So we want to develop some methods that use neural networks to filter the variables. Also, due to the large number of parameters of deep neural networks, it takes a lot of practice to train them, so it is also necessary to simplify the network structure when controlling the FDR.

In our methods, we use the $\ell_1$ regularisation loss function, determining the largest regularisation parameter \cite{taheri2021} that prevents the weight of the neural network from becoming zero. According to these parameters, integrating the integration method and neural network technology, we design three algorithms to screen variables. And based on the regularisation parameters corresponding to each weight, we select the weights whose parameters are relatively small and delete them so that the network can be simplified.

There are many application directions related to FDR control, and neural networks are also needed in many fields. Our research direction can be applied to many fields, and the following is a summary of the application areas based on the perspectives of some articles. For example, genomics research\cite{kassani2022},\cite{lin2012}, \cite{reimand2019} and \cite{sesia2019}. It can also be used in brain imaging studies, such as \cite{xuk2023} and \cite{Miller2016}.  For omics research, researchers in molecular biology and biomedical research often perform large-scale metabolomics, proteomics or other omics analyses \cite{Pittau2019}. It can also be used in particle physics experiments \cite{xie2021}, understanding the pathobiology of complex diseases \cite{abramovich2006} and discovering the association between the microbiome and obesity \cite{ithapu2017}.

\noindent\textbf{Outline of Paper}

\noindent 
The rest of this paper is organised as follows. 
Section \ref{sec:relatedworks} introduces related algorithms for filtering variables while controlling the FDR, and concludes the previous methods aimed at simplifying neural networks. Section \ref{sec:Methodology} details our methodologies, including using \textcolor{black}{one-layer neural networks, multiple-layer neural network}, and variable weight aggregation to filter the important variables. Section 
\ref{sec:Simulation} explains how our algorithms were simulated to verify their accuracy. Section 
\ref{sec:Application} applies our algorithms to a real dataset on breast cancer dataset and demonstrates their effectiveness in variable screening and neural network simplification. In Section 
\ref{sec:Discussion}, we discuss the advantages and disadvantages of our methods and outline possible extensions.

\section{Related Works}\label{sec:relatedworks}
The FDR played an important role in the selection of variables, providing a means to set constraints during variable filtering and serving as a metric to evaluate filtering performance. It represented the expected proportion of falsely selected variables among all the variables, where a false discovery was defined as a selected variable that did not actually appear in the true model. The FDR is defined as $$ FDR = E[\bm{V/(V+S)}], $$
\textcolor{black}{where $\bm{V}$ is the number of falsely selected variables, and $\bm{S}$ is the number of truly selected variables, such that $\bm{V+S}$ is the total number of selected variables. We define the ratio $\bm{V/(V+S)}$ to be $0$ when $\bm{V+S} = 0$.}

The control of FDR has had significant application in various research studies. Abramovich \cite{abramovich2006} introduced a data-adaptive thresholding scheme that ensured that only a certain expected fraction of rejected null hypotheses corresponded to false rejections by controlling the FDR. The method was asymptotically minimax for loss, simultaneously over a range of sparsity classes.
In addition, \cite{benjamini2009} proposed a novel FDR control procedure as a penalised model selection method. In \cite{benjamini2001}, a simple but effective procedure for FDR control was presented for independent test statistics. This procedure was shown to be superior to similar methods that traditionally control for family-wise error rates. 

Furthermore, the knockoffs method \cite{barber2015}, introduced by Barber and Candès in 2015, proved to be a powerful tool for multiple validation and feature selection, especially in the context of high-dimensional datasets. As the knockoffs method had a very good performance in filtering variables, it was widely used. DeepROCK \cite{Chen2023} used knockoffs and a novel DNN architecture to jointly control the false discovery rate and maximise statistical power. Numerous variants and extensions of knockoffs have been introduced, including those by \cite{Barber2019}, \cite{Candes2018}, and \cite{Romano2020}. In addition, \cite{jordon2018b} adapted the generative adversarial network framework to allow us to generate knockoffs without making assumptions about the feature distribution. 

Following the introduction of knockoff methods, several methods for variable selection using knockoffs or other nonlinear models have been proposed. Causal DAG-deepVASE (directed acyclic graphs using deep learning variable selection)\cite{fan2023} was the first computational method to explicitly learn nonlinear causal relationships and it estimated effect size using a deep neural network approach coupled with the knockoffs framework. DeepLINK \cite{zhu2021}, a sophisticated deep learning-based knockoffs inference framework, was developed to ensure robust false discovery rate control in high-dimensional scenarios. The DeepPINK model \cite{lu2018} was also a non-linear variable selection model.
In both articles\cite{Bai2020}, \cite{Ghosh2019}, their statistically efficient dimensionality reduction deep neural networks can and do perform well in variable filtering.  

To further improve the accuracy of the selected variables, some research on variable selection by ensemble learning methods has been proposed. Aggregating knockoffs \cite{xie2021} used the knockoffs method for $K$ times and combined the application results. 

In addition, a neural network was a powerful model that automatically learned the characteristics of the data and constantly updated the model parameters through an optimiser to minimise the loss function. This process was widely used in deep learning \cite{dziugaite2015}, especially in data filtering and model optimisation. \cite{dietterich2000} reviewed ensemble methods, explained why ensembles can often perform better than any single classifier. To ensure the rigour of the data screening, we can simulate many times and aggregate the results. This approach was similar to ensemble, which improved the robustness and generalisation of the model. By simulating many times, we can introduce diversity into the model training process, reducing the risk of overfitting and improving the model's performance on different datasets.

Due to the formidable capabilities of deep neural networks, there is currently a plethora of research dedicated to them, \cite{xuk2023} reviewed the existing research in this field. However, overfitting was a common problem in neural networks, and the use of large networks tended to be sluggish. Therefore, simplifying neural networks using appropriate methods was effective.  Dropout \cite{srivastava2014} was one trick to solve overfitting. However, because neurons need to be randomly dropped in each training step, dropout increased the training time, especially for large networks. Therefore, we want to develop a regular and more time-efficient method for simplifying neural networks.

\section{Methodology}\label{sec:Methodology}
\raggedbottom{In this section, we describe our four algorithms, which aim to select variables while controlling the FDR and simplifying the neural network.}

Our algorithms are suitable for data where there is a non-linear relationship between the independent variables and the response variables. We consider the $m$ independent variable vectors $\bm{x}^i=(x_1^i,\dots,x_p^i)^T\in\mathbb{R}^p$ and the dependent variable vectors $\bm{y}^i=(y_1^i,\dots,y_K^i)^T\in\mathbb{R}^K$ for $i=1,\dots,m$. 

Suppose that the set of variables truly related to the dependent variable is $S^* \subset \{1, \ldots, p\}$, and $|S^*| = s^* < p$. To enable rigorous false discovery control, we construct knockoff variables $\tilde{x}_j$ for each original variable $x_j$, following the framework proposed by Barber and Candès \cite{barber2015}.

The idea is to generate knockoff variables that share the same distributional properties as the original variables, particularly their pairwise correlations, while ensuring that they are conditionally independent of the response. Formally, let $\textcolor{black}{\bm{x}} \in \mathbb{R}^p$ denote a random feature vector from the underlying population with covariance matrix $\Sigma = \text{Cov}(\textcolor{black}{\bm{x}}) \in \mathbb{R}^{p \times p}$. We construct a random knockoff feature vector $\tilde{\textcolor{black}{\bm{x}}} \in \mathbb{R}^p$ such that:
\begin{equation*}
    \text{Cov}(\textcolor{black}{\bm{x}}) = \text{Cov}(\tilde{\textcolor{black}{\bm{x}}}) = \Sigma, \quad \text{Cov}(\textcolor{black}{\bm{x}, \tilde{\bm{x}}}) = \Sigma - \text{diag}(s),
\end{equation*}
where $s \in \mathbb{R}^p$ is a non-negative vector. A commonly used choice is to set $s_j = 2\lambda_{\min}(\Sigma) \wedge 1$ for all $j$, where $\lambda_{\min}(\Sigma)$ denotes the smallest eigenvalue of $\Sigma$. This choice yields the so-called equi-correlated knockoffs and ensures valid construction when $\Sigma$ is positive definite.

In practical implementation, we assume access to the data matrix $X \in \mathbb{R}^{\textcolor{black}{m} \times p}$ whose rows are the observed feature vectors. When $\textcolor{black}{m} \geq 2p$, the knockoff variables can be constructed as:
\[
\tilde{X} = X (I - \Sigma^{-1} \mathrm{diag}(s)) + U C,
\]
where $U \in \mathbb{R}^{\textcolor{black}{m} \times p}$ is an orthonormal matrix orthogonal to the column space of $X$, and $C \in \mathbb{R}^{p \times p}$ satisfies $C^\top C = 2\mathrm{diag}(s) - \mathrm{diag}(s) \Sigma^{-1} \mathrm{diag}(s)$. This ensures that the augmented design matrix $[X, \tilde{X}]$ satisfies the desired joint covariance structure.

Thus, for each original sample $\bm{x}^i$, its corresponding knockoff feature vector is $\tilde{\bm x}^i = (\tilde{x}_1^i, \ldots, \tilde{x}_p^i)^\top$. These constructed knockoff variables serve as negative controls, and their comparison with the original variables forms the basis for valid variable selection and false discovery rate control.

Simultaneously, we have designed a neural network with \(L\) hidden layers. The input layer has a total of \(2p\) neurons (representing \(p\) independent variables and $p$ knockoffs variables of them). The $l^{th}$ layer has \(h_l\) neurons (\(0 \leq l \leq L+1\)), where $h_0 = 2p$ means the number of input neurons, and the output layer consists of $h_{L+1} = K$ neurons. Let $\bm{\Theta}$ be the \textcolor{black}{list of parameter matrices}, where $\bm{\Theta} = (\bm{\theta}^{(1)}, \ldots, \bm{\theta}^{(L+1)})$. Here, $\bm{\theta}^{(l)} \in \mathbb{R}^{h_l \times h_{l-1}}$ for $l = 1,\ldots,L+1$, and $\theta_{jd}^{(l)}$ denotes the weight connecting the $d^{\text{th}}$ neuron in layer $l-1$ to the $j^{\text{th}}$ neuron in layer $l$.
  
\begin{figure}[h]
    \centering
    \includegraphics[width=0.53\linewidth]{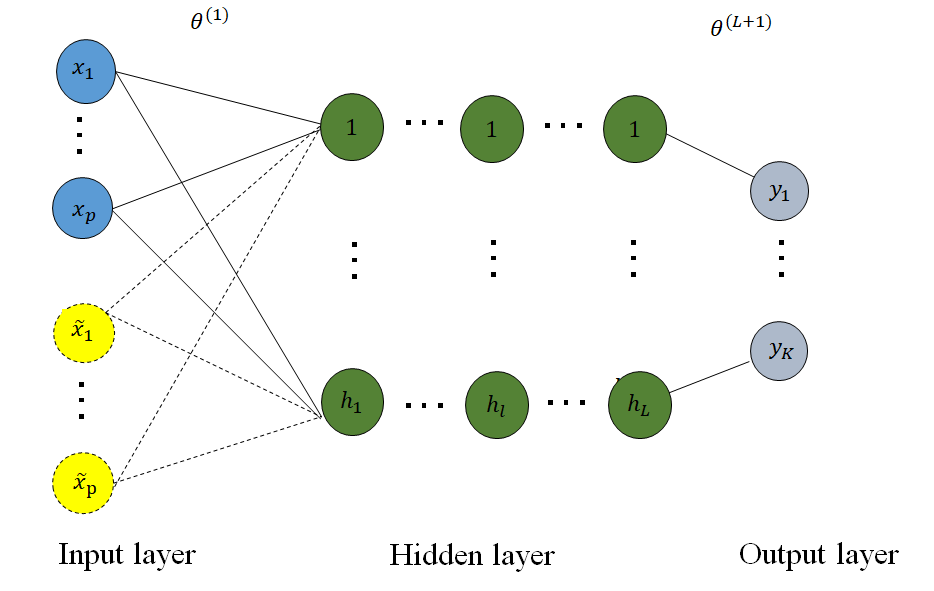}
    \caption{The figure shows an example of a neural network. $x_1,\ldots,x_p,\widetilde{x}_1,\ldots,\widetilde{x}_p$ are the inputs, and $y_1,\dots,y_K$ are the outputs. This neural network has a total of $L$ hidden layers, and $\bm{\theta}^{(1)},\ldots, \bm{\theta}^{(L+1)}$ are the weight matrices of the network. \textcolor{black}{The numerical labels in the hidden layers indicate neuron indices.}}
    \label{fig1}
\end{figure}

Let $\bm{u}^i = ((\bm{x}^i)^T, (\widetilde{\bm{x}}^i)^T)^T\in\mathbb{R}^{2p}$ represent the values of each variable and its corresponding knockoffs variable for the $i^{th}$ \textcolor{black}{sample}. The output of the neural network for the $i^{th}$ example can be written as 
\[
h_{\bm{\Theta}}(\bm{u}^{i}) = \bm{\sigma}_{L+1}\left(\bm{\theta}^{(L+1)}\left(\bm{\sigma}_{L}\left(\bm{\theta}^{(L)}\ldots\bm{\sigma}_{1}\left(\bm{\theta}^{(1)}\bm{u}^i\right)\right)\right)\right), \quad i=1,\dots,m,
\]
where $\bm{\sigma}_1, \ldots,\bm{\sigma}_{L+1}$ are the entrywise activation functions.

In order to use the neural network to successfully filter out the independent variables associated with the response variable, while controlling the FDR and simplifying our network, we should obtain the importance parameters of the weights, which can be obtained from the regularised loss function.

\begin{align*}
 J(\bm{\Theta}) = &\frac{1}{m}\sum_{i=1}^{m}\sum_{k=1}^{K}\left[-y_k^{i}\log((h_{\bm{\Theta}}(\bm{u}^{i}))_k) - (1-y_k^{i})\log(1-(h_{\bm{\Theta}}(\bm{u}^{i}))_k) \right]\\
 &+\frac{\lambda}{2m}\sum_{l=1}^{L+1}\sum_{j=1}^{h_{l-1}}\sum_{d=1}^{h_{l}}\left| \theta _{jd}^{(l)}\right|, 
\end{align*}
where $\lambda>0$ is a regularisation parameter. 

Note that the above loss function is based on a binary cross-entropy formulation, which is appropriate for multi-label binary classification tasks, where each label $y_k^i \in \{0,1\}$. In cases where the response $\bm{y}^i$ is a one-hot vector corresponding to a multi-class classification problem, our framework remains applicable by replacing the sigmoid activation and binary loss with a softmax activation and categorical cross-entropy loss. In fact, such a setup is used in our experiments (Section \ref{sec:Simulation}) where the outcome variable involves more than two classes.

And we define FDR as

$$ FDR = E\left[ \frac{\# \{j:j \notin S^* \textnormal{ and } j \in \widehat{S}\}}{\# \{j:j \in \widehat{S}\} \vee 1 }\right], $$
where $S^*$ is a set of variables that are truly related to the dependent variable and $\widehat{S}$ is the set of selected variables. \textcolor{black}{Note that this set-theoretic definition is mathematically identical to the classical definition $FDR = E[V/(V+S)]$ introduced in Section 2. Here, the numerator corresponds to the number of false discoveries ($V$), and the denominator corresponds to the total number of selected variables ($V+S$). The $\vee 1$ operator (taking the maximum with 1) handles the boundary case where $\widehat{S}$ is empty, ensuring the ratio evaluates to 0.}

How do we get $\widehat{S}$? We know that when using regularised gradient descent, if the weight is relatively important, reducing it to zero requires a larger corresponding regularisation parameter. In addition, each variable in the input layer connects to multiple neurons in the next layer through different weights. If the connected weights are important (i.e., remain large after regularisation), then the input variable is considered important to the model prediction. Therefore, we can use this idea for filtering variables using neural networks.

And proximal gradient descent combines gradient descent with the proximal operator to effectively handle non-smooth regularisation terms. By using the proximal operator of the \( L1 \) regularisation, it induces sparse solutions, i.e. it sets small weights to zero. Therefore, we use proximal gradient descent in our algorithm to generate sparse weights for our model.

\noindent
A summary of our algorithms is given below: 

\noindent
\textbf{One Layer Filter}: screens variables based only on the regularisation parameters of each weight in the first layer of the net.

\noindent
\textbf{Multiple Layer Filter}: combines the regularisation parameters of the weights in the first layer with all the weights in the neural network to filter variables.

\noindent
\textbf{Variable Weight Aggregation Filter}: trains the model many times, uses hard voting to filter variables, and gets the aggregated variables.

\noindent
\textbf{Reduce Weight}: based on the regularisation parameters of each weight obtained during variable screening to simplify the structure of neural networks.

\subsection{One Layer Filter}\label{3.1}
When training the model, experiment with different values of $\lambda$ for each weight in the network, and find the largest regularisation parameter that prevents the weight from becoming zero. The larger the parameter value, the more important the weight. Variables connected to more important weights are also more important.
To determine whether a parameter truly remains nonzero under a given $\lambda$, we train the neural network until convergence based on the validation loss. Specifically, we use an early stopping criterion that monitors the validation loss for a fixed number of epochs (e.g., 10) without improvement. This ensures that parameters which remain nonzero are stable across training. If no such criterion is used, a parameter may appear nonzero prematurely but eventually shrink to zero in later epochs.

We acknowledge that retraining the network for each value of $\lambda$ can be computationally expensive, especially when fine-grained $\lambda$ grids are explored. In our implementation, training for all $\lambda$ values can be parallelised and performed in batches. 

\begin{algorithm}[h]
\caption{One Layer Filter}\label{algorithm1}
\SetKwInOut{Input}{Input}
\SetKwInOut{Output}{Output}
\Input{samples $\bm{x}^i$, knockoffs $\widetilde{\bm{x}}^i$, label $y^i$, $i=1,\dots,m$, neural network model with $L$ hidden layers, target FDR $q$,
$\lambda_{\textnormal{min}} =\textnormal{min}\{Z_{jd}
^{(l)}:1 \le l \le L+1, 1 \le j \le h_{l-1}, 1 \le d \le h_l\}$, $\lambda_{\textnormal{max}} =\textnormal{max}\{Z_{jd}
^{(l)}:1 \le l \le L+1, 1 \le j \le h_{l-1}, 1 \le d\le h_l\} $.}

\Output{selected variables set $\widehat{S}$.}
\For{$\lambda = \lambda_{\min}$ \KwTo $\lambda_{\max}$}{
    \If{$\theta_{jd}^{(l)} == 0 $}{
        $Z_{jd}^{(l)} = \lambda$
    }
} 
 Compute importance of variable $j$ and \textcolor{black}{its} corresponding knockoffs variable:
   $z_j =\sum_{d=1}^{h_1}Z_{jd}^{(1)}, \quad \widetilde{z}_j = \sum_{d=1}^{h_1}Z_{(j+p)d}^{(1)}, \quad 1 \le j \le p.$ \\

  Compute test statistic for variable: $$W_{j} = z_{j} \vee \widetilde{z}_{j} \cdot 
\Big\{
\begin{array}{ll}
    +1, & \textrm{if } z_{j} \ge \widetilde{z}_{j}, \\
    -1, & \textrm{if } z_{j} < \widetilde{z}_{j}.
\end{array}$$ \\
$\mathcal{W} = \{|W_j|: j = 1, \ldots , p\}$ \\

Compute data-dependent threshold for $q$:
    $T = \min\{ t \in \mathcal{W} : \frac{\#\{j:W_{j}\le -t\}}{\#\{j:W_{j}\ge t\} \vee 1} \le q \}$.
  
The selected model is  $\widehat{S}=\{j:W_j \ge T\}$.
\end{algorithm}

Feed the variables $x_1,\ldots,x_p, \widetilde{x}_1,\ldots,\widetilde{x}_p$ into a neural network model with $L$ hidden layers. Training the model with the standard optimisation algorithm and the regularisation loss function defined earlier, we can find the
\textcolor{black}{
\begin{equation*}
Z_{jd}^{(l)} = \sup\{\lambda : \theta_{jd}^{(l)}(\lambda) \neq 0\}, \quad \text{for } l = 1, \dots, L+1
\end{equation*}
}
is the largest $\lambda$ that prevents each weight in the model from becoming zero. To simplify the notation, we will omit the $\lambda$ of $\theta_{jd}^{(l)}(\lambda)$ in the following. 
The importance for the variable $j$ and its knockoffs can be defined as $$z_j =\sum_{d=1}^{h_1}Z_{jd}^{(1)}, \quad \widetilde{z}_j = \sum_{d=1}^{h_1}Z_{(j+p)d}^{(1)}, \quad 1 \le j \le p,$$ where $h_1$ is the number of neurons in the first hidden layer. Set 
\[
W_{j} = z_{j} \vee \widetilde{z}_{j} \cdot
\left\{
\begin{array}{ll}
    +1, & \textrm{if } z_{j} \ge \widetilde{z}_{j}, \\
    -1, & \textrm{if } z_{j} < \widetilde{z}_{j}.
\end{array}
\right.
\]    
A data-dependent threshold for the target FDR $q$ is 
\[
T = \min\left\{ t \in \mathcal{W} : \frac{\#\{j:W_{j}\le -t\}}{\#\{j:W_{j}\ge t\} \vee 1} \le q \right\},
\]
where $\mathcal{W} = \{|W_j|: j = 1, \ldots , p\}$. Then, the selected model is defined as $\widehat{S}=\{j:W_j \ge T\}$.\\

Explanation of a single layer filter:
When adjusting the size of the regularisation coefficient during model training, the values of each weight $\theta_{jd}^{(l)}$ in the neural network also change after training. If $\lambda$ is large enough, $\theta_{jd}^{(l)}$ will approach zero after training. Each weight $\theta_{jd}^{(l)}$ has a corresponding $\lambda$ that makes it zero, and this $\lambda$ is relatively larger, indicating that the weight $\theta_{jd}^{(l)}$ is relatively important, and vice versa. Therefore, ${Z_{jd}^{(l)}}$ can be seen as a representation of the importance of the weight $\theta_{jd}^{(l)}$. 
In the neural network, a variable in the input layer is connected to different weights. If it is connected to more important weights (with larger ${Z_{jd}^{(1)}}$), it means that this variable is relatively important. So we can use $z_j$ or $\widetilde{z}_{j}$ to represent the importance of a variable. A larger total $z_j$ indicates greater importance for that variable. Similarly to linear knockoffs, a large positive value of $W_j$ indicates that the variable $x_j$ enters the model early and before its knockoffs copy $\widetilde {x}_j$.
Therefore, this is an indication that this variable is a true signal. We want to select variables for which $W_j$ is large and positive, i.e. $W_j \geq T$ for some $T \ge 0$. And the fraction appearing above the data-dependent threshold $T$ is an estimate of the proportion of false discoveries when we select all features $j$ with $W_j \geq T$.

\subsection{Multiple Layers Filter}\label{3.2}
It's worth noting that if we only use one layer, we can't take the whole neural network into account and some information may be lost. Therefore, we should pay attention to the other layers as well. This is why we introduce the multiple layers filter.
\begin{algorithm}[h]
\caption{Multiple Layers Filter}\label{algorithm2}
\SetKwInOut{Input}{Input}
\SetKwInOut{Output}{Output}
\Input{samples $\bm{x}^i$, knockoffs $\widetilde{\bm{x}}^i$, label $y^i$, $i=1,\dots,m$, neural network model with $L$ hidden layers, target FDR $q$,
$\lambda_{\textnormal{min}} =\textnormal{min}\{Z_{jd}
^{(l)}:1 \le l \le L+1, 1 \le j \le h_{l-1}, 1 \le d \le h_l\}$, $\lambda_{\textnormal{max}} =\textnormal{max}\{Z_{jd}
^{(l)}:1 \le l \le L+1, 1 \le j \le h_{l-1}, 1 \le d \le h_l\} $.}

\Output{selected variables set $\widehat{S}$.}
\For{$\lambda = \lambda_{\min}$ \KwTo $\lambda_{\max}$}{
    \If{$\theta_{jd}^{(l)} == 0$}{
        $Z_{jd}^{(l)} = \lambda$
    }
}
Calculate the product sum of the weight matrix of each layer:

\If{Exist one of $\bm{\theta}^{(l)}$ is closed to $0$  ($\le {10}^{-4}$)}{use Min-Max Normalization to the 
$\bm{\theta}^{(l)}$}
\If{Binary classification}{
$\bm{w} = {\left( \bm{\theta}^{(L+1)} \times\cdots\times\bm{\theta}^{(1)} \right)^\top}$}
\If{Multi-class classification}{
$\bm{o} = {\left( \bm{\theta}^{(L+1)} \times\cdots\times\bm{\theta}^{(1)} \right)^\top}$,
$w_j = \sum_{k=1}^{K}o_{jk},1 \le j \le {2p}$
}

Compute importance of variable $j$ and its corresponding knockoffs variable:
    $z_j = \sum_{i=1}^{h_1}Z_{jd}^{(1)}, \widetilde{z}_j = \sum_{i=1}^{h_1}\widetilde{Z}_{ji}^{(1)},$
    $g_j = z_j \cdot w_j$ and $\widetilde{g}_j = \widetilde{z}_j \cdot {w}_{j+p}$.

 Compute statistic test for variable $j$: $W_j = g_j^2 - \widetilde{g}_j^2$\\
 $\mathcal{W} = \{|W_j|: j = 1, \ldots , p\}$ \\

Compute data-dependent threshold for $q$:
    $T = \min\{ t \in \mathcal{W} : \frac{\#\{j:W_{j}\le -t\}}{\#\{j:W_{j}\ge t\} \vee 1} \le q \}$. \\
  
Get the selected model $\widehat{S}=\{j:W_j \ge T\}$.
\end{algorithm}

\textcolor{black}{For binary classification problems, let $\bm{w} = \left( \bm{\theta}^{(L+1)} \times\cdots\times\bm{\theta}^{(1)} \right)^\top\in\mathbb{R}^{2p}$, where $\bm{w}$ represents the global weight vector. }
\textcolor{black}{
For multi-class classification problems, set $\bm{o} = \left( \bm{\theta}^{(L+1)} \times\cdots\times\bm{\theta}^{(1)} \right)^\top\in\mathbb{R}^{2p\times K}$, and $w_j = \sum_{k=1}^{K} o_{jk}$ for $1 \le j \le 2p$.}

In this method, when calculating $\bm{w}$, we use weights with suitable penalties for the product of the weight matrices. 
Set $W_j = g_j^2 - \widetilde{g}_j^2$, where $g_j = z_j \cdot w_j$ and $\widetilde{g}_j = \widetilde{z}_j \cdot {w}_{\color{red}j+p}$.
The data-dependent threshold is then determined by
\begin{equation*}
    T = \min\left\{ t \in \mathcal{W} : \frac{\#\{j:W_{j}\le -t\}}{\#\{j:W_{j}\ge t\} \vee 1} \le q \right\}.
\end{equation*}
Then, the selected model is $\widehat{S}=\{j:\textit{W}_j\ge T\}$.

Explanation of the multi-layer filter: The weight $\theta_{jd}^{(l)}$ carries important information about the neural network. By integrating the weight, we can include information from the whole neural network. So we can use $\bm{w}$ to include the whole neural network. Then we use $g_j = z_j \cdot w_j$ to represent the importance of each variable and $\widetilde{g}_j = \widetilde{z}_j \cdot {w}_{\color{red}j+p}$ to represent the importance of the knockoffs variable corresponding to each variable, if $g_j$ is greater, $\widetilde{g}_j$ is smaller, $W_j$ = $g_j^2 - \widetilde{g}_j^2$ is greater. Since $W_j$ represents the difference between the importance of the original variable and its corresponding knockoffs variable, it can also be defined in other ways, such as $W_j =|g_j| - |\widetilde{g}_j|$. While a large positive value of $W_j$ indicates that the variable $x_j$ enters the model earlier than its knockoff $\widetilde{x}_j$. This is an indication that the variable is a true signal. Our goal is to select variables for which $W_j$ is significantly large, and the fraction above $T$ serves as an estimate of the proportion of false positives when we select all features $j$ with $W_j \geq T$. 

At first glance, this approach seems similar to DeepPINK \cite{lu2018}, but overall there are notable differences. First, in terms of neural network structure, our method does not include a filter layer. Secondly, the combination of weights in each layer is different. Finally, the methods for defining the importance of variables are also different.

\subsection{Variable Weight Aggregation Filter}\label{3.3}

In \ref{3.2}, we screened the variables through one layer filter and multiple layers filter. Here, we used multiple model fitting to obtain the average FDR, which can not only control FDR well but also reduce the influence of the randomness of neural network. In order to facilitate stable and interpretable feature selection, the Variable Weight Aggregation (VWA) Filter method performs multiple runs of model training on a fixed dataset and aggregates the selected variables across runs. This approach reduces the randomness introduced by neural network training and highlights variables that are consistently selected, thereby improving the stability and reliability of the feature selection results.

\begin{algorithm}[H]
\caption{VWA Filter}\label{algorithm3}
\SetKwInOut{Input}{Input}
\SetKwInOut{Output}{Output}
\SetKwInOut{Initialization}{Initialization}

\Input{Number of repeated model trainings $n$, filter result $\widehat{S}^{(a)}$ from Algorithm \ref{algorithm1} or Algorithm \ref{algorithm2} in the $a$-th simulation, the number of all the variables $p$, the number of occurrences of variable $j$ in the $n$ simulations $count[j]$, the proportion of filtered variables in the number of simulations $r$}
\Output{Aggregated set $\widehat{S}_{A}$}
\Initialization{$\widehat{S}_{A}$ = \{\}; $count[j]$ = 0, $j = 1,\cdots, p$}

\BlankLine

Calculate the number of times the variable occurs in each simulation:

\For{$a = 1$ \KwTo $n$}{
    \For{$j \in \widehat{S}^{(a)}$}{
        $count[j] = count[j] + 1$}
    }
Filter out the selected variables:\\
\For{$j = 1$ \KwTo $p$}{
    \If{$count[j]/n >= r$}{
        $\widehat{S}_{A} = \widehat{S}_{A}
        \cup \{j\}$
    }
}
Get the selected model $\widehat{S}_{A}$.

\end{algorithm}

In Algorithm \ref{algorithm3}, we use the idea of aggregation and the voting method to ensure that the final variable is filtered several times in multiple fits. So we can reduce errors and thus control FDR. This algorithm can be used for aggregation of one layer filter, multiple layers filter, or both these two. Here, $n$ denotes the number of repeated model fittings used to estimate variable selection stability, and we will compare the results by adjusting $n$ later. We recommend choosing 0.5 for $r$ here. Choosing 0.5 as the threshold value for the frequency of feature occurrence is a reasonable compromise to ensure the consistency and importance of the selected features in multiple models, and also to improve the stability and interpretability of the model. In practical applications, this threshold can be appropriately adjusted according to specific needs and the characteristics of the dataset, and the stringency of the screening can be changed by adjusting the proportion of the occurrence of the filtered variables.

\textcolor{black}{Note that the $n$ repeated models share the exact same architecture and hyperparameters. The variations in selected variables across different iterations stem purely from the inherent stochasticity of neural network training (e.g., random weight initialization and data shuffling). By aggregating results across $n$ independent runs, our VWA filter mitigates this randomness to identify the most stable features.}

\subsection{Reduce Weight}\label{3.4}
 




In variable filtering, we can record the $Z_{jd}^{(l)}$ values, which indicate the importance of the corresponding weights $\theta_{jd}^{(l)}$. Based on the information provided by $Z_{jd}^{(l)}$, we can delete some less important weights. Set $c\in[0,1]$ as the deletion rate for model weights, ${D} = \{\theta_{jd}^{(l)} : Z_{jd}^{(l)} \le e_c\}$ are the weights to delete, where $$e_c = \max\bigg\{a: \frac{\#\{Z_{jd}^{(l)} : Z_{jd}^{(l)} \le a, \text{for\ all\ } j, d, l\}}{\#\{Z_{jd}^{(l)} : \text{all\ } j, d, l\}} \le c\bigg\},$$
and ${D}^C$ are the remaining weights. Deleting all ownership values in the previous layer that point to a particular neuron is the same as deleting the entire neuron.

For our neural network, we first reduce the input neurons by variable selection. We then further simplified the neural network by using weight reduction to reduce both the weights and the neurons.


\section{Simulation}\label{sec:Simulation}

To validate the accuracy of our algorithms, we generate different datasets to simulate the processes of variable selection and neural network simplification. Throughout the simulation, we create datasets with different sample sizes and distributions, ensuring that the relationships between the independent and dependent variables in these datasets are non-linear. Specifically, the $p$ input variables in $\bm{x} = (x_1, \ldots, x_p)^T$ are mutually independent and sampled from either a standard normal distribution $\mathcal{N}(0, 1)$ or a chi-square distribution $\chi^2(4)$ with 4 degrees of freedom, depending on the simulation setting. The associated knockoff variables $\tilde{\bm{x}}$ are generated following the procedure in [3]. For each simulation, we define a subset $S^* \subset \{1, \ldots, p\}$ as the set of truly relevant variables, and set $|S^*| = s^*$. The dependent variable $y$ is generated by applying a non-linear transformation (a data-generating neural network) to a linear combination of the selected relevant variables $x_j$ for $j \in S^*$, with fixed coefficients. We optionally add Gaussian noise to modulate the signal strength, enabling control over the difficulty of the variable selection task. \textcolor{black}{Please note that the architecture of this data-generating neural network is structurally different from the network used later for variable selection, reflecting a realistic scenario where the true underlying model is unknown.} 

{In our setting, each dataset has a total of $p$ variables $\bm{x} = (x_{1}, \ldots, x_{p})^T$, and the variables related to $y$ are $\bm{x}_r$. Our dependent variable $y$ is given by}

\begin{equation*}
{y^i = \sigma_{L+1} \left( \bm{\theta}^{(L+1)} \left( \sigma_L(\cdots \sigma_1(\bm{\theta}^{(1)} \bm{x}^i) \cdots ) \right) \right), \quad i=1,\dots,n,}
\end{equation*}
where $\sigma_1, \ldots, \sigma_{L}$ are the ReLU activation functions. For binary classification, $\sigma_{\color{red}L+1}$ is the step function. The threshold of the step function is a value that makes the positive and negative examples equally distributed. For multi-classification, $\sigma_{\color{red}L+1}$ is the softmax function.

In this section, we evaluate the performance of the three filters by observing the filtering results of different numbers of variables, variable distributions, and neural network structures. In order to better evaluate the performance of our method, we also define the power, which represents the ability of the models to select variables related to the dependent variables. We define it as 
$$ Power = E\left[ \frac{\# \{j:j \in S^* \textnormal{ and } j \in \widehat{S}\}}{\# \{j:j \in S^*\} \vee 1 }\right] .$$

The following analysis focuses on two parts: the results of different filters and the results of reduced weight methods. We also compare our methods with existing methods.

\subsection{Comparison of Selection Results}\label{4.1}

\textbf{One Layer Filter and Multiple Layers Filter:} 

\begin{figure}[h]
    \centering
    \includegraphics[width=1\linewidth]{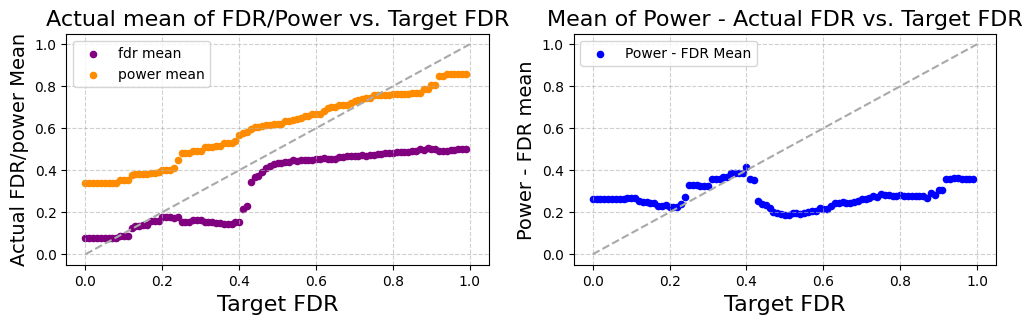}
    \caption{This figure illustrates the variation in Power, FDR, and Power-FDR for a one layer filter when different values of $q$ are set.}
    \vspace{-2mm}
    \label{fig2}
\end{figure}

\begin{figure}[h]
    \centering
    \includegraphics[width=1\linewidth]{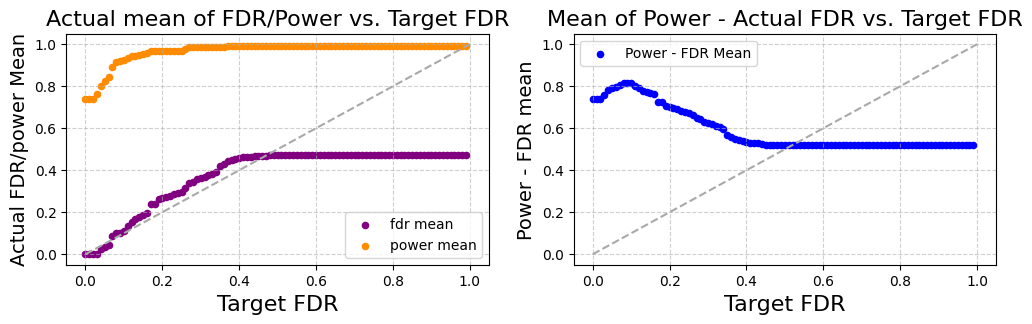}
    \caption{This figure illustrates the variation in Power, FDR, and Power-FDR for a multiple layers filter when different values of target FDR $q$} are set.
    \vspace{-2mm}
    \label{fig3}
\end{figure}

We repeated the training process multiple times under the same experimental settings (e.g., fixed model architecture, data distribution, and q values) to estimate the mean and variability of FDR and Power. By varying the number of experiments from 5 to 100, we found that the mean and variance of FDR, Power and Power-FDR remained relatively stable, indicating that our model is robust. Figure \ref{fig2} and Figure \ref{fig3} show the average results from ten experiments. The graph on the left shows how the actual FDR changes as the target FDR $q$ increases. The right-hand chart illustrates how the difference between power and actual FDR changes as $q$ increases. Through
this graph, we can identify the point with the best performance of the model, which
is characterized by the largest difference between power and actual FDR. For Figure \ref{fig2}, \textcolor{black}{the largest Power--FDR difference for the one layer filter occurs when \(q=0.4\).} For Figure \ref{fig3}, \textcolor{black}{the largest Power--FDR difference for the multiple layers filter occurs when the target FDR \(q\) is around 0.1.} \textcolor{black}{These results indicate that the empirical FDR and Power are sensitive to the nominal target level \(q\). In the following, we fix \(q=0.1\) and further examine the corresponding knockoff statistics and selected variables.}

\begin{figure}[h]
    \centering
    \begin{minipage}{0.48\linewidth}
        \centering
        \includegraphics[width=\linewidth]{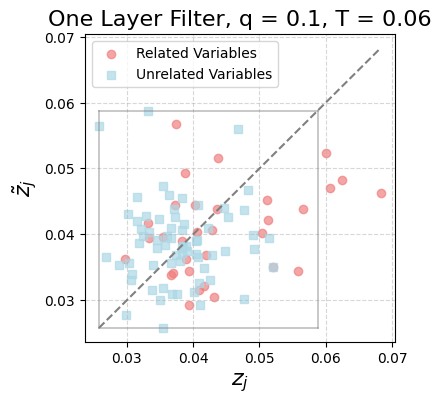}
        \caption{This figure illustrates the filtering results of one layer filter, when $q$=0.1.}
        \vspace{-2mm}
        \label{fig4}
    \end{minipage}
    \hfill
    \begin{minipage}{0.5\linewidth}
        \centering
        \includegraphics[width=\linewidth]{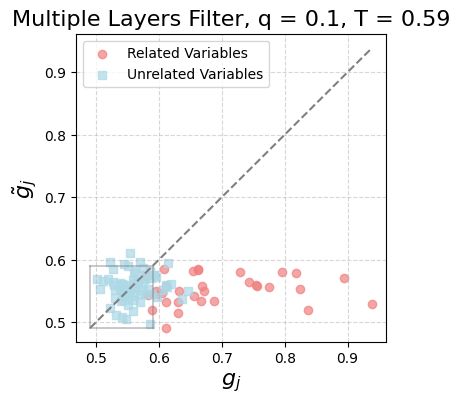}
        \caption{This figure illustrates the filtering results of multiple layers filter, when $q$=0.1.}
        \vspace{-2mm}
        \label{fig5}
    \end{minipage}
\end{figure}

\textcolor{black}{With \(q=0.1\), we obtain the filtering results of the one layer filter and the multiple layers filter.} In Figure \ref{fig4}, the x-axis represents $z_j$ and the y-axis represents $\widetilde z_j$. And in Figure \ref{fig5}, the x-axis represents $g_j$ and the y-axis represents $\widetilde g_j$. Square dots represent unrelated variables and circular dots represent relevant variables. The number of points to the right of the vertical line and below the diagonal equals $\#{\{j: W_j \ge T}\}$, representing the selected variables when the data-dependent threshold $T$. The number of points above the horizontal line and to the left of the diagonal corresponds to $\#{\{j: W_j \leq -T}\}$. Note that the true signals are mainly below the diagonal, indicating $W_j \ge 0$. The nulls are roughly symmetrically distributed along the diagonal. As $q$ increases, $T$ moves to the left, leading to the selection of more and more variables.

\textbf{VWA Filter:} 
During the course of our experiments, we noticed that there were slight variations in the variables derived from each training. Based on this observation, and in order to reduce the impact of this randomness on our FDR and power, we decided to combine each result in a summarized manner. We tried different model training durations: 10, 20, and 50 times. After 10 training sessions, we compared the aggregated FDR and power with the results from 20 and 50 sessions. The results are almost identical, indicating that increasing the training duration does not significantly improve performance and that 10 training sessions are sufficient to improve the experimental results. Considering the increased cost of training a large model multiple times, we chose to conduct the simulation experiment with 10 training sessions.

In the following discussion, we use the following conventions to make this easier to understand: OL: One Layer Filter,
ML: Multiple Layers Filter,
VWA-OL: VWA (One Layer) Filter,
VWA-ML: VWA (Multiple Layers) Filter,
VWA-OML: VWA (One and Multiple Layers) Filter.

We consider the various scenarios with different settings, and the filtered results are shown in Table~\ref{tab1}. There are three types of neural network model structures: $(200,264,1)$, $(20,10,1)$, $(200,264,128,1)$. Two types of data distributions were used: a chi-square distribution and a normal distribution. For the total number of variables and the number of related variables, we set several cases as shown in Table \ref{tab1}.

Here we attempt to aggregate the fitting results of one layer filter 
and multiple layers filter separately, \textcolor{black}{and evaluate 
their performance in terms of both empirical FDR and Power,} where 
the variable filtering ratio is set to 0.5. We try to aggregate the results of the two models together to get the VWA-OML model, where the variable filtering ratio is set to 0.25. This model reduces the variable proportion criterion. It ensures that even when the VWA-OL and VWA-ML results are poor, variables that may be underestimated in some cases can be found, thus improving the overall discovery rate and validity. As a result, when the VWA-OL and VWA-ML results perform well, VWA-OML can be opted out and adjusted flexibly based on the performance of other models. By integrating the results of one layer filter and multiple layers filter, the VWA-OML model is able to utilize the advantages of both to ensure that reliable features can still be obtained when the results of different models are inconsistent.

\textcolor{black}{As shown in Table~\ref{tab1} and Table~\ref{tab1.2}, some VWA results have empirical FDR values close to or equal to zero. We agree that a very low empirical FDR should not be interpreted as evidence of superior performance by itself, because it may also reflect overly conservative variable selection. Therefore, we evaluate the proposed methods jointly in terms of empirical FDR and Power.}

\textcolor{black}{The existing simulation results indicate that the VWA procedures are not simply avoiding false discoveries by selecting almost no variables. For example, in the binary classification settings in Table~\ref{tab1}, VWA-OL achieves empirical FDR values close to zero while maintaining Power values of 0.61, 1.00, 0.95, 0.79, 0.80, and 0.45 across different settings. Similarly, VWA-OML achieves near-zero empirical FDR with Power values of 0.85, 1.00, 1.00, 0.97, 1.00, and 0.48. A similar pattern is observed in the multi-class classification settings in Table~\ref{tab1.2}, where several VWA variants also maintain non-trivial Power together with low empirical FDR. These results suggest that the low empirical FDR values are not merely due to selecting no variables; the methods still recover a substantial proportion of truly relevant variables in many settings.}

\textcolor{black}{At the same time, some variants can be more conservative. For example, VWA-ML has lower Power in some settings, indicating that the aggregation step may remove weak but relevant variables when they are not selected consistently across repeated neural-network trainings. Hence, the practical performance of the proposed methods should be assessed through the FDR--Power trade-off rather than by FDR alone.}

In addition, we simulate a three-class classification problem to examine whether the proposed methods can be extended beyond binary classification. Table~\ref{tab1.2} reports the corresponding multi-class selection results under normal and chi-square feature distributions. To further examine the sensitivity of the empirical FDR and Power to network architecture, feature distribution, and the proportion of relevant variables, we also provide supplementary simulation results in Appendix~\ref{sec:appendix}, Tables~\ref{tab:normal_binary}--\ref{tab:gamma_muti}.

\textcolor{black}{Taken together, the simulation results suggest that the proposed methods achieve reasonable empirical FDR--Power trade-offs in the normal settings and in some chi-square settings. However, the supplementary Gamma-distributed settings show that the empirical FDR can exceed the nominal target level \(q=0.1\), especially under skewed non-Gaussian feature distributions. Therefore, these findings should be interpreted as empirical evidence of distribution-dependent variable	screening performance, rather than as a guarantee of exact finite-sample FDR control for arbitrary feature distributions.}

\vspace{-0.1em}

\begin{table}[h]
\footnotesize
\resizebox{\textwidth}{!}{%
\small
\begin{tabular}{|c|c|c|c|c|c|c|c|c|}
\hline
\multicolumn{3}{|c|}{Model structure} & \multicolumn{4}{c|}{(200,264,1)} & (20,10,1) & (200,264,128,1) \\
\hline
\multicolumn{3}{|c|}{Distribution} & chi-square distribution (df=4) & \multicolumn{5}{c|}{Normal distribution} \\
\hline
\multicolumn{3}{|c|}{Total variables} & 100 variables & \multicolumn{3}{c|}{100 variables} & 10 variables & 100 variables \\
\hline
\multicolumn{3}{|c|}{Related variables} & 33 & 10 & 20 & 33 & 5 & 33 \\
\hline
\multirow{2}{*}{Filters} & \multirow{2}{*}{OL} & FDR & 0.12(0.11)& 0.09(0.88)& 0.03(0.20)& 0.00(0.04)& 0.05(0.30)& 0.05(0.01)\\
\cline{3-9}
& &  Power & 0.34(0.01)& 0.32(0.29)& 0.50(0.19)& 0.43(0.09)& 0.6(0.10)& 0.22(0.33)\\
\cline{2-9}
& \multirow{2}{*}{ML} & FDR & 0.22(0.10)& 0.14(0.32)& 0.11(0.02)& 0.15(0.03)& 0.00(0.33)& 0.03(0.00)\\
\cline{3-9}
& &  Power & 0.11(0.01) & 0.66(0.05)& 0.78(0.06)& 0.94(0.28)& 0.80(0.24)& 0.39(0.02)\\
\cline{2-9}
& \multirow{2}{*}{VWA-OL} & FDR & 0.00(0.00) & 0.00(0.07) & 0.00(0.02) & 0.00(0.00) & 0.00(0.00) & 0.00(0.07) \\
\cline{3-9}
& &  Power & 0.61(0.03) & 1.00(0.00) & 0.95(0.04) & 0.79(0.07)& 0.80(0.13) & 0.45(0.07) \\
\cline{2-9}
& \multirow{2}{*}{VWA-ML} & FDR & 0.00(0.00) & 0.00(0.00) & 0.00(0.00) & 0.00(0.02)& 0.00(0.00) & 0.00(0.00) \\
\cline{3-9}
& &  Power & 0.42(0.06) & 0.80(0.05) & 0.40(0.10) & 0.48(0.10) & 0.40(0.18) & 0.12(0.05) \\
\cline{2-9}
& \multirow{2}{*}{VWA-OML} & FDR & 0.00(0.04) & 0.00(0.12) & 0.00(0.08) & 0.00(0.07) & 0.00(0.07) & 0.00(0.08) \\
\cline{3-9}
& &  Power & 0.85(0.09) & 1.00(0.00) & 1.00(0.00) & 0.97(0.09)& 1.00(0.00) & 0.48(0.27) \\
\hline
\end{tabular}%
}

    \centering
    \vspace{0.6em}
    \caption{Filtered results for binary classification when $q$ = 0.1. The number of experiments is 10, and the corresponding standard deviation is shown in parentheses.}
    \label{tab1}
\end{table}

\vspace{-1em}
\begin{table}[h]
\footnotesize
\resizebox{\textwidth}{!}{%
\small
\begin{tabular}{|c|c|c|c|c|c|c|c|c|}
\hline
\multicolumn{3}{|c|}{Model structure} & \multicolumn{4}{c|}{(200,264,3)} & (20,10,3) & (200,264,128,3) \\
\hline
\multicolumn{3}{|c|}{Distribution} & chi-square distribution (df=4) & \multicolumn{5}{c|}{Normal distribution} \\
\hline
\multicolumn{3}{|c|}{Total variables} & 100 variables & \multicolumn{3}{c|}{100 variables} & 10 variables & 100 variables \\
\hline
\multicolumn{3}{|c|}{Related variables} & 33 & 10 & 20 & 33 & 5 & 33 \\
\hline
\multirow{2}{*}{Filters} & \multirow{2}{*}{OL} & FDR & 0.21(0.21)&  0.00(0.21)&  0.11(0.07)& 0.10(0.02)& 0.00(0.40)& 0.06(0.40)\\
\cline{3-9}
& &  Power & 0.30(0.05)& 0.30(0.01)& 0.25(0.01)& 0.33(0.30)&0.4(0.29)& 0.33(0.18)\\
\cline{2-9}
& \multirow{2}{*}{ML} & FDR &0.10(0.22)&0.01(0.30)& 0.03(0.23)&0.09(0.20)&  0.06(0.18)& 0.12(0.20)\\
\cline{3-9}
& &  Power & 0.20(0.33)& 0.80(0.27)&0.59(0.10)&0.76(0.17)&0.78(0.16)& 0.42(0.25)\\
\cline{2-9}
& \multirow{2}{*}{VWA-OL} & FDR &0.00(0.00)  &0.29(0.11)  &0.00(0.04)  &0.11(0.06)  & 0.00(0.00) &0.00(0.09)  \\
\cline{3-9}
& &  Power &1.00(0.10)  &1.00(0.14)  &1.00(0.00)  &1.00(0.16) &0.80(0.14)  &0.70(0.04) \\
\cline{2-9}
& \multirow{2}{*}{VWA-ML} & FDR &0.00(0.00)  &0.00(0.12)  &0.00(0.00)  &0.00(0.02)  &0.00(0.00)  &0.00(0.25)  \\
\cline{3-9}
& &  Power &1.00(0.16)  &0.70(0.10)  &1.00(0.02)  &1.00(0.10)  &0.40(0.18)  &0.12(0.09)  \\
\cline{2-9}
& \multirow{2}{*}{VWA-OML} & FDR &0.20(0.06)  &0.29(0.21)  &0.17(0.19)  &0.13(0.07)  & 0.00(0.09) &0.00(0.11)  \\
\cline{3-9}
& &  Power &1.00(0.00)  &1.00(0.00)  &1.00(0.00)  &1.00(0.00)  &1.00(0.00)  &0.70(0.22)  \\
\hline
\end{tabular}%
}
    \centering
    \vspace{0.6em}
    \caption{Filtered results for multi-class classification when $q$ = 0.1. The number of experiments is 10, and the corresponding standard deviation is shown in parentheses.}
    \label{tab1.2}
\end{table}

\subsection{Reduce Weight}\label{4.2}
According to the values of $Z_{jd}^{(l)}$ obtained from variable filtering, as mentioned in the paper, we eliminate weights with relatively small $Z_{jd}^{(l)}$. In this section, we illustrate the weight structure of the neural network after weight reduction for data sets with only 10 variables, where 5 variables are related to $y$. Here we set the weight deletion ratio to 0.5. In the Appendix \ref{sec:appendix}, Figures \ref{fig9} and \ref{fig10} show the weight distribution of the model after weight reduction. 
In Figures \ref{fig9} and \ref{fig10}, if there is no connection between neurons, it means that the weight between them has been removed. If there are no weight connections either before or after a hidden neuron, the hidden neuron is deleted and marked with a dashed line. 

We also use the filtered results from different filters for prediction.

\begin{table}[h]
\footnotesize

\begin{center}
    \begin{tabular}[h]{|p{1.1cm}p{1.9cm}p{1.08cm}p{1.08cm}p{1.4cm}p{1.4cm}p{1.55cm}p{1.9cm}|}  
    \hline
    & & OL & ML & VWA-OL &VWA-ML&VWA-OML& All variables\\
    \hline
    Time(s) & General & 10.43 & 10.43 & 10.42 & 10.42& 10.84&11.21 \\
    & Dropout & 11.92 & 12.03 & 10.99 & 10.81& 13.29& 16.67\\
    & Reduce weight & 9.33 & 9.49 & 10 & 9.46& 9.70& 11.25\\
    \hline
    Accuracy & General & 78.25$\%$ & 77.45$\%$ & 70.65$\%$ & 75.2$\%$& 80.35$\%$&75.85$\%$\\
    & Dropout & 77.60$\%$ & 72.05$\%$ & 67.40$\%$& 70.35$\%$& 80.05$\%$ & 76.70$\%$\\
    & Reduce weight & 78.11$\%$& 77.60$\%$& 71.20$\%$& 76.90$\%$& 79.90$\%$& 76.71$\%$\\
    \hline
    \end{tabular}%
    \end{center}
    \centering

    \caption{The prediction results of different algorithms on 100 variables with 33 variables are related to the dependent variable $y$. The ``ALL variables" column represents the results of the prediction using all variables.}
    \label{tab2}
    \end{table}

Throughout the process, we conduct a comparative analysis of the average experimental outcomes between the reduced weight and dropout methods and the approach that involves no processing on the network. All variables in the table are the predicted time and results for all variables that have not been filtered. This comparison is based on 25 experiments, where we calculate the average values of accuracy and time. The results are summarized in Table \ref{tab2}. Compared to dropout, our approach uses less time than dropout, but is more accurate than dropout. Compared to General, which has no simplified model, our method takes less time and our algorithm accuracy is almost higher than general. Therefore, our approach demonstrates the commendable performance of our algorithm in variable filtering and streamlining neural network structures to reduce operational time.

\subsection{Comparison with Existing Technology}\label{4.3}
To better examine the performance of each of our algorithms, we also compare our algorithms internally to each other and to existing techniques (knockoffs filter for linear model \cite{barber2015}, DeepLINK \cite{zhu2021}, DeepPINK \cite{lu2018}, DFS (Deep Feature Selection) \cite{chen2021} and DeepLasso \cite{Dinh2020}). For these methods, the data we use are binary normally distributed data, where there are 100 variables and 33 variables are related to the dependent variable. For the imitation filters, DeepLINK, DeepPINK and DFS, we set $q$ to 0.1. The performance of the different models has its own advantages. Knockoffs perform well when there is a linear relationship between features and target variables, but can be limited when dealing with complex data. DeepLINK and DeepPINK excel at capturing complex non-linear relationships, especially in high-dimensional data and complex feature spaces. In addition, DFS and DeepLasso do not aim to control FDR, their variable screening strategies are different from ours, and our model can be better evaluated by comparing the results. \textcolor{black}{The detailed comparison results are reported in Table~\ref{tab3}.}

\begin{table}[h]
\footnotesize
\begin{center}
    \begin{tabular}{|p{3cm}|p{7cm}|p{1cm}|p{1cm}|}
        \hline
        Method & Variable & FDR &  Power \\
        \hline
       Knockoffs & $x_{14}, x_{16}, x_{28}, x_{32}$ & 0 & 0.12 \\
        \hline
         VWA-OL & \
        $x_5, x_6, x_8, x_9, x_{12}, x_{13}, x_{14}, x_{17}, x_{23}, x_{26}, x_{27}, x_{29}, x_{19}$, & 0 & 0.79 \\
        &$x_{32}, x_{1}, x_{22}, x_{15}, x_{31}, x_{2}, x_7, x_{33}, x_{20}, x_{21}, x_3, x_{4}, x_{30}$&&\\
        \hline

         VWA-ML & 
        $x_{14}, x_{27}, x_{29}, x_{12}, x_{17},x_{23}, x_6, x_7, x_{25}, x_{28},x_{32}, x_{10}, x_{11}$, & 0 & 0.48 \\
        &$x_{19},x_{31}, x_{16}$&&\\
        \hline
        
        VWA-OML & 
        $x_5, x_6, x_8, x_9, x_{12}, x_{13}, x_{14}, x_{17}, x_{23}, x_{26}, x_{27}, x_{29}, x_{19}$, & 0 & 0.97 \\
        &$x_{32}, x_{1}, x_{22}, x_{15}, x_{31}, x_{2}, x_7, x_{33}, x_{20}, x_{21}, x_3, x_{4}, x_{30}$,&&\\
        &$x_{10}, x_{18}, x_{25}, x_{16}, x_{11}, x_{28}$&&\\
        \hline
        
        DeepLINK(ko) & 
        $x_{11}, x_{16}, x_{21}, x_{24}$ & 0 & 0.12 \\
        \hline

        DeepPINK & 
        $x_{28}, x_{44}, x_{56}$ & 0.67 & 0.03 \\
        \hline
        
        DFS &
        $x_1, x_9, x_{17}, x_{20}, x_{23}, x_{25} ,x_{26}, x_{27}, x_{30}, x_{31}$ & 0 & 0.30 \\
        
        \hline
        DeepLasso &
        First 5 variables: $x_{23},x_{27},x_{10},x_{26},x_{6}$& \textbackslash & \textbackslash \\
        \hline
    
    \end{tabular}
    \end{center}
    
    \centering
    \caption{The comparison of our algorithms and those of existing algorithms, the data has 100 features, and the first 33 variables are related to dependent variable $y$. For the first four methods, we set the $q$ = 0.1.}
    \label{tab3}
\end{table}

\textbf{One Layer Filter:} This is a relatively simple filter that uses only one layer of the neural network. While it does not consider the entire model structure, it has proven effective in screening out relevant variables during experiments.

\textbf{Multiple Layers Filter:} An extension of the one layer filter, this method takes into account the entire structure of the neural network, giving better results for variable filtering.

\textbf{VWA Filter:} VWA is a method that aggregates feature selection results from one layer filter and multiple layers filter to improve discovery rates and validity, allowing for flexible adaptability based on model performance while reducing the influence of training randomness.

\textbf{Reduce Weight:} We delete the less important weight according to the weight importance obtained during the selection of the variables and achieve the simplification of the neural network. Through experiments, we find that this method can simplify the network and reduce the training time of the model while maintaining accuracy. 

\section{Application} \label{sec:Application}
The Breast Cancer Wisconsin (Diagnostic) \text{Dataset}\textsuperscript{\tiny{\ref{dataset}}} (WDBC) contains attributes calculated from a digitised image of a fine needle aspirate (FNA) of a breast mass. It is primarily used for classification tasks, with real-valued attributes derived from FNAs of breast masses, with a total of 569 instances and 30 features.
\footnotetext[1]{\label{dataset}Wolberg, W., Mangasarian, O., Street, N., \& Street, W. (1995). Breast Cancer Wisconsin (Diagnostic). UCI Machine Learning Repository. \href{http://dx.doi.org/10.24432/C5DW2B}{http://dx.doi.org/10.24432/C5DW2B}}


The label in this dataset is `Diagnosis', which indicates the outcome of the medical diagnosis of breast cancer. \textcolor{black}{The two diagnostic labels are summarized in Table~\ref{tab4}.}

\begin{table}[h]
\footnotesize
\begin{center}
\begin{tabular}{|p{2.5cm}|p{10.5cm}|}
  \hline
  \textbf{Label} & \textbf{Description} \\
  \hline
  M = Malignant & Indicates the presence of cancerous cells. \\
  B = Benign & Indicates the absence of cancerous cells, typically considered harmless. \\
  \hline
  
\end{tabular}%
\end{center}
\centering
\caption{Label and description of Breast Cancer Wisconsin (Diagnostic) dataset}
\label{tab4}
\end{table}

\textcolor{black}{The features are computed from digitized images of fine needle aspirates (FNA) of breast masses and describe characteristics of the cell nuclei present in each image. The ten basic nuclear morphology features are described in Table~\ref{tab5}. For each basic feature, three statistical summaries are provided: the mean, the standard error, and the worst/largest value. Therefore, the 30 predictors do not correspond to measurements from three different nuclei; instead, they correspond to three statistical summaries of the ten nuclear morphology features.}

\begin{table}[h]
\footnotesize
\begin{center}
\begin{tabular}{|p{2.5cm}|p{10.5cm}|}
  \hline
  \textbf{Feature} & \textbf{Explanation} \\
  \hline
  Radius & Measure the average distance from the center of the nucleus to points on its perimeter. \\
  Texture & Denote the standard deviation of gray-scale values within the nucleus. \\
  Perimeter & Account for the total distance between contiguous points forming the nuclear boundary. \\
  Area & Give the total pixel count within the nucleus boundary in the digital image. \\
  Smoothness & Evaluate local variations in radius lengths. \\
  Compactness & Calculated as \(\frac{\textnormal{Perimeter}^2}{\textnormal{Area}} - $1.0$ \). \\
  Concavity & Gauge the severity of concave portions along the nuclear contour. \\
  Concave Points & Count the number of concave portions or ``indents" along the contour. \\
  Symmetry & Assess the symmetry of the cell nucleus. \\
  Fractal Dimension & Derived as ``coastline approximation" minus 1. \\
  \hline
\end{tabular}%
\end{center}
\centering
\centering
\caption{Features and description}
\label{tab5}
\end{table}

\textcolor{black}{For the convenience of notation, we denote the 30 predictors by $x_1,\ldots,x_{30}$, as shown in Table~\ref{tab6}. The suffixes 1, 2, and 3 in the original variable names do not indicate the first, second, and third nuclei. Instead, they correspond to three statistical summaries: mean, standard error, and worst/largest value, respectively. For example, radius1, radius2, and radius3 correspond to mean radius, radius standard error, and worst radius, respectively.}

\begin{table}[h]
\footnotesize
\begin{center}
\begin{tabular}{|l|l|l|l|l|l|}
\hline
Column Name & $\bm{x}$ & Column Name &  $\bm{x}$& Column Name &  $\bm{x}$ \\
\hline
radius1 & \(x_1\) & texture1 & \(x_2\) & perimeter1 & \(x_3\) \\
area1 & \(x_4\) & smoothness1 & \(x_5\) & compactness1 & \(x_6\) \\
concavity1 & \(x_7\) & concave\_points1 & \(x_8\) & symmetry1 & \(x_9\) \\
fractal\_dimension1 & \(x_{10}\) & radius2 & \(x_{11}\) & texture2 & \(x_{12}\) \\
perimeter2 & \(x_{13}\) & area2 & \(x_{14}\) & smoothness2 & \(x_{15}\) \\
compactness2 & \(x_{16}\) & concavity2 & \(x_{17}\) & concave\_points2 & \(x_{18}\) \\
symmetry2 & \(x_{19}\) & fractal\_dimension2 & \(x_{20}\) & radius3 & \(x_{21}\) \\
texture3 & \(x_{22}\) & perimeter3 & \(x_{23}\) & area3 & \(x_{24}\) \\
smoothness3 & \(x_{25}\) & compactness3 & \(x_{26}\) & concavity3 & \(x_{27}\) \\
concave\_points3 & \(x_{28}\) & symmetry3 & \(x_{29}\) & fractal\_dimension3 & \(x_{30}\) \\
\hline
\end{tabular}%
\end{center}
\centering
\caption{Feature mapping for breast cancer data}
\label{tab6}
\end{table}

\subsection{Optimizers for Aggregation}\label{5.1}
Here we use the method of filtering variables. From Table \ref{tab7}, we can see the different results of filters. In Section \ref{5.2} we will predict the results of each filter and compare the results. It can be seen that VWA-OML can screen out more variables because the screening ratio is not so strict, and whether the screened variables are effective depends on the predicted results.
\begin{table}[h]
\footnotesize
    \begin{center}
    \begin{tabular}{|p{2cm}|p{11cm}|p{1cm}|}
 \hline
 \textbf{Filter} & \textbf{The selected variables} & \textbf{Count} \\
 \hline
 OL & $x_1,x_2,x_3,x_8,x_9, x_{10},x_{15}, x_{18},x_{25},x_{26}, x_{27}, x_{29}, x_{30}$& 13\\
 ML & $x_1,x_2,x_8,x_{10},x_{12}, x_{15},x_{17},x_{25},x_{26},x_{28}$ & 10 \\
 VWA-OL & $x_{2}, x_{15}, x_{18}, x_{19}, x_{21}, x_{23}, x_{28}$ & 7\\
 VWA-ML & $x_3, x_{15}$ & 2\\
 VWA-OML & $x_{2}, x_{15}, x_{18}, x_{19}, x_{21}, x_{23}, x_{28}, x_3, x_7, x_{10}, x_{25}, x_{29}, x_{30}, x_{27}, x_8, x_9, x_{20}, x_{24}, x_{26}, x_5, x_4$ & 21 \\
 \hline
\end{tabular}
\end{center}
	\centering
	\caption{Screening results of three filters for breast cancer data, $q =0.01$}
	\label{tab7}
\end{table}

\subsection{Reduce Weight}\label{5.2}
 
Our breast cancer prediction models are structured as $(\#\{x_{\textnormal{selected}}\} \times 10 \times 1)$, where \(\#\{x_{\textnormal{selected}}\}\) is the number of variables selected for each filter when \(q=0.01\). We set the weight deletion ratio to 0.3. The number of neurons removed from each layer is recorded in the Appendix \ref{sec:appendix}, Table \ref{tab8}, with neurons represented as \((n_1, \ldots, n_{10})\) for the hidden layer.

\begin{figure}[h]
\centering\includegraphics[width=0.7\linewidth]{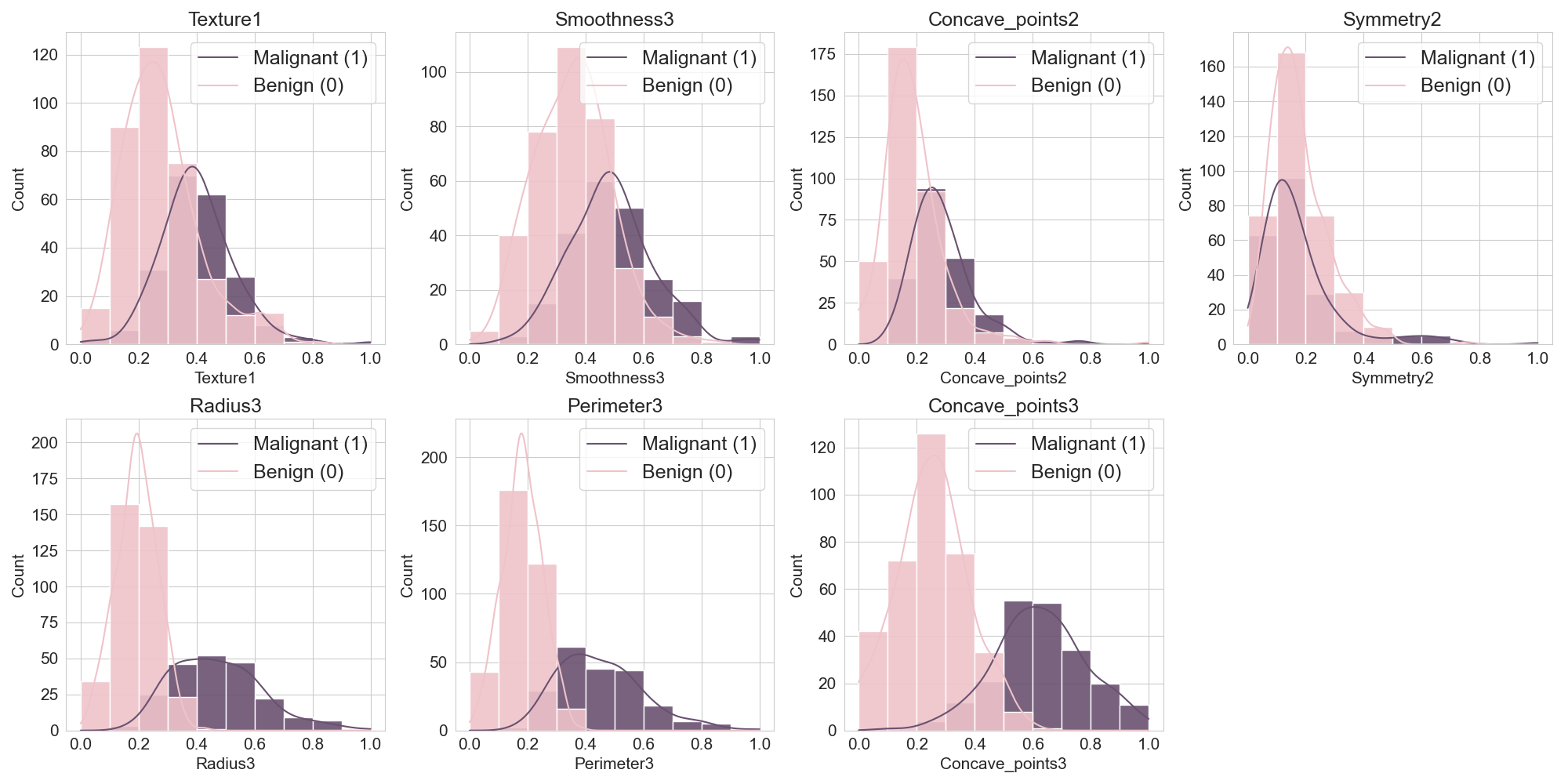}
    \caption{Distributions of the variables selected by VWA-OL. The selected variables are interpreted according to the WDBC feature summaries in Table \ref{tab6}.}
    \label{fig12}
\end{figure}

We use the results of different filters for prediction. During the prediction process, we explore two simplified neural network methods and perform a comparative analysis of the average accuracy and time for each experimental result, as shown in the Appendix~\ref{sec:appendix}, Table~\ref{tab9}. It can also be found from Table~\ref{tab9} that our method takes less time than dropout and achieves comparable or higher accuracy. \textcolor{black}{The distributions of the variables selected by VWA-OL are shown in Figure~\ref{fig12}.}

Here we can see that in general the most accurate filter VWA-OL result is screened for variables that include: $x_2$, $x_{15}$, $x_{18}$, $x_{19}$, $x_{21}$, $x_{23}$, $x_{28}$. The results showed that seven characteristics of the data in the diseased cells were related to whether or not cancer had developed. Our results are also consistent with the research of most scientists \cite{Pienta1991}. 
A total of seven core features are associated with dependent variables: texture1, smoothness2, concave points2, symmetry2, radius3, perimeter3, concave points3. We can see that the distributions of each feature are different for the benign and malignant results.

\subsection{Comparison with Existing Technology}\label{5.4}

We also experiment with other methods on this dataset 
and compare them with ours. Since the true relevant variables are 
unknown in this real-world dataset, we use prediction accuracy as 
the evaluation metric rather than FDR. We tested the three VWA 
variants (VWA-OL, VWA-ML, and VWA-OML) under the same setting and 
obtained average accuracies of 98.25\%, 79.82\%, and 97.02\%, 
respectively. As VWA-OL achieved the highest accuracy, we selected 
it for comparison with other existing methods.

\textcolor{black}{We note that the better performance of VWA-OL in 
this case study does not necessarily contradict the simulation 
results. In the simulations, the methods are evaluated by empirical 
FDR and Power under known ground truth, whereas in the WDBC 
real-data application, prediction accuracy is used instead. 
Moreover, the WDBC dataset has only 569 samples and may differ 
from the controlled simulations in signal structure, feature 
correlation, and noise level. Since VWA-ML uses information 
propagated through multiple layers, it may be more sensitive to 
model complexity or training instability in this small-sample 
setting. Therefore, the relative performance of VWA-OL and VWA-ML 
should be understood as dataset-dependent rather than as a general 
contradiction of the simulation findings.}

From Table \ref{tab10} we can see that the prediction accuracy of our method is higher than the other five methods.

\begin{table}[h]
\footnotesize
    \begin{center}
    \begin{tabular}{|p{4cm}|p{7cm}|p{2cm}|}
        \hline
        Method & Variable & Accuracy \\
        \hline
        Knockoffs & $x_{8}, x_{11}, x_{25}, x_{28}$ & 93.33\% \\
        \hline
        
        VWA-OL &
        $x_{2}, x_{15}, x_{18}, x_{19}, x_{21}, x_{23}, x_{28}$ & 98.25\% \\
        
        \hline
        
        DeepLINK(ko) & $x_1, x_2, x_6, x_7, x_8,x_{17}, x_{23}, x_{24}$& 95.52\% \\
        
        \hline

        DeepPINK & $x_{21}, x_{28}$ & 95.61\%\\

        \hline
         DFS & $x_3, x_{23}$ & 95.61\%\\
        
        \hline
        
        DeepLasso & First 5 variables: $x_{17}, x_{32}, x_{31}, x_{30},x_{29}$ & 95.66\%\\
        
        \hline

    \end{tabular}%
    \end{center}
    \centering
    \caption{Comparison of our algorithm's variable selection results and prediction accuracy with those of existing algorithms. And for the first four methods we set $q$ = 0.01.}
    \label{tab10}
\end{table}

\section{Discussion}\label{sec:Discussion}
Based on knockoff algorithms and the loss function of neural networks, this paper designs three algorithms to filter variables with neural networks under the condition of controlling the occurrence of errors. The one layer filter is simpler and faster in variable selection than the other two methods, since it only requires the calculation of the parameters of the first layer of the neural network. However, a disadvantage is that it doesn't use all the information from the neural network, which can lead to a loss of information. To overcome this limitation, we introduced the multiple layers filter, which is more comprehensive than the one layer filter approach. However, it is more complex to implement than the one layer filter. The VWA approach can combine two filters that can adjust the proportion and number of variables screened to more interpretable variables by voting, as well as reduce the proportion of variable occurrences to screen out variables that may have been overlooked.

To simplify the network, the weights of the neural network are reduced, and even neurons of the neural network are deleted, according to the importance parameters of the weights of the neural network obtained in the screening process, to simplify the structure of the neural network. The algorithm we have developed is remarkably systematic. We ran simulations on self-created datasets and applied our algorithm to real datasets, such as the Wisconsin Breast Cancer dataset. We compared our three methods with the original knockoffs method in terms of variable screening. We also compared our weight reduction method with dropout when simplifying the architecture of deep neural networks. Our results show that the VWA filter excels in variable filtering, and the weight reduction method is particularly effective in simplifying deep neural networks while maintaining a high level of prediction accuracy. Consequently, we believe that our work is well suited for identifying factors associated with specific variables and for reducing the training time of deep neural network models.

However, there are some shortcomings in our model: First, the independent variables of the data should preferably be continuous. Second, the neural network model is inherently more complex than a simple linear model. Consequently, the time required for variable filtering is longer in the neural network model. Third, when simplifying the neural network, it is critical to set the weight deletion ratio judiciously. If the deletion ratio is too high, this will lead to underfitting, which can affect the effectiveness of the model. Fourth, deep neural networks, characterised by their multiple layers and large number of parameters, are better suited to large data sets. The increased complexity of such networks allows them to learn and represent complex functions effectively, provided the dataset is large enough to prevent overfitting. At last, in our current implementation, input variables are assumed to be metric. This may limit applicability in medical datasets that include categorical or ordinal variables, such as sex or clinical stages. Although our method currently does not incorporate such features, we acknowledge that these variables could be included via standard preprocessing methods (e.g., one-hot encoding), and we plan to explore this extension in future work. \textcolor{black}{Another limitation is that the empirical FDR performance depends on the feature distribution and on the quality of the knockoff construction. In particular, the Gamma-distributed simulations show that the empirical FDR can exceed the nominal target level when the features are strongly right-skewed and non-Gaussian. This suggests that the current second-order knockoff construction may be insufficient for exact FDR control in all non-Gaussian settings. Future work will consider more flexible knockoff generators, such as model-X knockoffs or deep generative knockoff constructions, to improve FDR control under skewed and complex feature distributions.}

Future directions for this research include optimising algorithms for improved speed and scalability, potentially integrating additional features such as gene expression data or patient history to improve model performance, extending validation to diverse datasets from different demographics or medical institutions to assess generalisability, conducting clinical trials to validate practical utility if predictive accuracy is maintained, adapting the methodology to diagnose and classify other diseases, and developing methods to balance underfitting and overfitting for improved model adaptability. These efforts have the potential to significantly advance medical diagnostics, machine learning and data analytics. The optimisation algorithm makes our methods applicable to a wider variety of datasets, including data and images with different distributions.

\section*{Acknowledgement(s)}

This work was supported in part by the Guangdong Basic and Applied Basic Research Foundation (No.~2023A1515110469),  the Guangdong Provincial Key Laboratory IRADS (No. 2022B1212010006), the grant of Higher Education Enhancement Plan (No. 2025KTSCX186), and the National Natural Science Foundation of China (No. 12271047). The authors would like to thank Huiqi Zhang and Xiaobo Huang for their help with the revision of this paper.

\section*{Code}

The code for this paper is available on \href{https://github.com/FangXieLab/Knockoff-DNN/tree/main}{Github}.

\newpage
\appendix 
\section{}
\label{sec:appendix}

\textbf{Simulation: Results of Reduce Weight (Section \ref{4.2})}
\begin{figure}[h]
    \centering
    \includegraphics[width=0.5\linewidth]{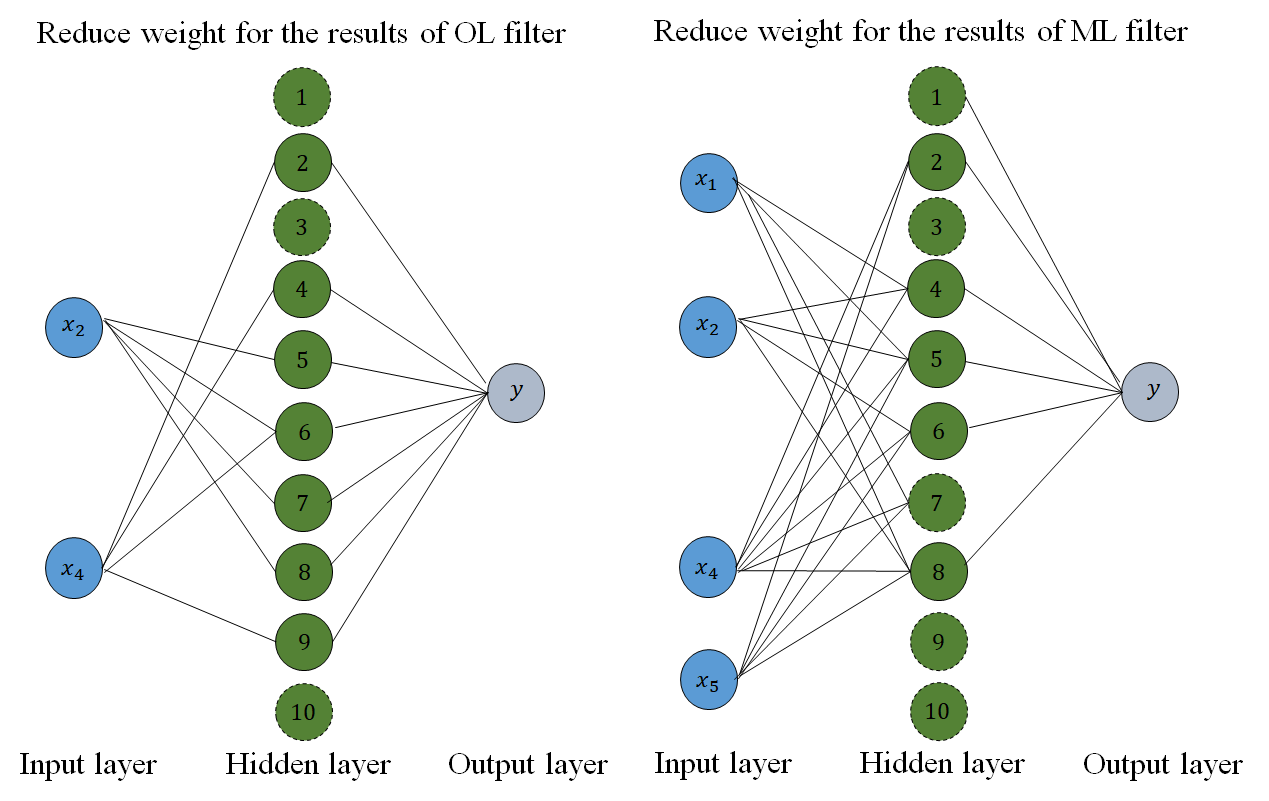}
    \vspace{-5mm} 
    \caption{Reduce weight results for OL and ML filter when deletion rate $c$ = 0.5}
    \label{fig9}
\end{figure}

\begin{figure}[h]
    \centering
    \includegraphics[width=0.5\linewidth]{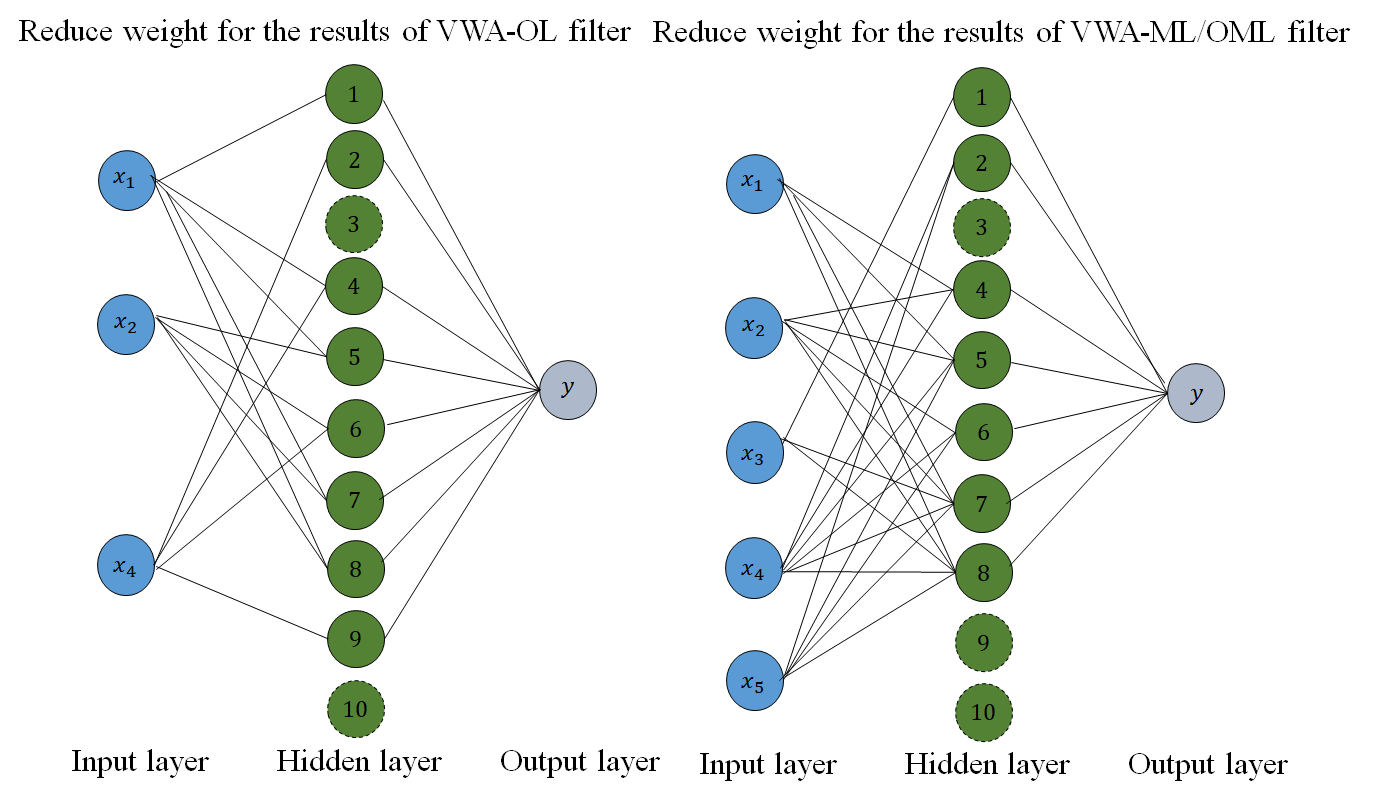}
    \vspace{-5mm} 
    \caption{Reduce weight results for VWA filter when deletion rate $c$ = 0.5}
    \label{fig10}
\end{figure}

\noindent
\textbf{Application: Results of Reduce Weight (Section \ref{5.2})}

For example, in \ref{3.1}, the value corresponding to the first layer and $h_7$ is 13, indicating that 13 weights connecting input neurons to the $7^{th}$ hidden neuron have been removed. Additionally, we observed that a one layer filter screened a total of 13 variables (see Table \ref{tab8}). Consequently, the connections between the input neurons and the $7^{th}$ hidden neuron were entirely removed, equivalent to the removal of the  $7^{th}$ hidden neuron.\\
\begin{table}[h]
\footnotesize
    \begin{center}
      \begin{tabular}{|l|l|*{10}{l}|l|}
    \hline
    
    & &$n_1$ & $n_2$  & $n_3$  & $n_4$ & $n_5$  & $n_6$  & $n_7$  & $n_8$  & $n_9$ & $n_{10}$  \\
    \hline
    OL& First layer&  1 & 1 & 2 & 4 & 2 & 4 & 13 & 13 & 0 & 2 \\
     & Second layer & 0 & 0 & 0 & 0 & 0 & 0 & 0 & 0 & 0 & 0 \\
    \hline
    ML& First layer& 0 & 3 & 0 & 3 & 0 & 0 & 3 & 3 & 0 & 0 \\
    & Second layer&0 & 0 & 0 & 0 & 0 & 0 & 0 & 0 & 0 & 0 \\
    \hline
    VWA-OL& First layer&  0 & 3 & 0 & 3 & 0 & 0 & 3 & 3 & 0 & 0 \\
   & Second layer&  0 & 0 & 0 & 0 & 0 & 0 & 0 & 0 & 0 & 0 \\
    \hline
    VWA-ML& First layer&  1 & 1 & 2 & 1 & 1 & 12 & 12 & 4 & 4 & 1 \\
    & Second layer & 0 & 0 & 0 & 0 & 0 & 0 & 0 & 0 & 0 & 0 \\
    \hline
    VWA-ML& First layer & 1 & 1 & 2 & 1 & 7 & 6 & 16 & 16 & 0 & 1 \\
   & Second layer  &0 & 0 & 0 & 0 & 0 & 0 & 0 & 0 & 0 & 0 \\
    \hline
      \end{tabular}%
    \end{center}
    \centering
    \caption{The structure of the different algorithms after reducing weight for breast cancer data}
\label{tab8}
\end{table}

\noindent
\textbf{Application: Results of Prediction (Section \ref{5.2})}

\begin{table}[h]
\footnotesize
\begin{center}
\begin{tabular}{|p{1.4cm}p{2.3cm}p{1.1cm}p{1.1cm}p{1.6cm}p{1.6cm}p{1.9cm}|} 
\hline
&& OL & ML & VWA-OL& VWA-ML& VWA-OML\\
\hline
Time(s) & General& 7.15& 8.13& 7.54& 7.61&7.55\\

&Dropout &7.42& 9.41& 8.09& 8.34& 8.16\\

&Reduce weight &6.33& 5.45& 7& 6.73& 7.66\\
\hline
Accuracy& General & 97.67$\%$& 97.73$\%$& 97.63$\%$&  83.33$\%$&   97.46$\%$\\

&Dropout&97.10$\%$& 95.08$\%$&  97.02$\%$&  82.81$\%$&  97.19$\%$\\

&Reduce weight& 97.64$\%$&96.55$\%$&  98.25$\%$& 79.82$\%$& 97.10$\%$\\
\hline
\end{tabular}%
\end{center}
\centering
\caption{The prediction results of different algorithms}
\label{tab9}
\end{table}

\newpage
\noindent
\textbf{Simulation: Results of Simulation (Section \ref{4.1})}

\begin{table}[h]
\scriptsize
\centering
\setlength{\tabcolsep}{6pt}
\renewcommand{\arraystretch}{1.1}
\resizebox{\textwidth}{!}{  
\begin{tabular}{|c|c||*{9}{c|}}
\hline
\multicolumn{2}{|c||}{Model structure} & \multicolumn{3}{c|}{(50,64,1)} & \multicolumn{3}{c|}{(100,264,1)} & \multicolumn{3}{c|}{(300,264,1)} \\
\hline
\multicolumn{2}{|c||}{Distribution} & \multicolumn{9}{c|}{Normal distribution, mean = 0, std = 1} \\
\hline
\multicolumn{2}{|c||}{Total variables} & \multicolumn{3}{c|}{25 variables} & \multicolumn{3}{c|}{50 variables} & \multicolumn{3}{c|}{150 variables} \\
\hline
\multicolumn{2}{|c||}{Related variables} & 5 & 10 & 20 & 10 & 25 & 40 & 100 & 150 & 200 \\
\hline
OL & FDR   & 0.07(0.01) & 0.05(0.04) & 0.02(0.05) & 0.07(0.04) & 0.07(0.02) & 0.08(0.01) & 0.03(0.04) & 0.09(0.04) & 0.04(0.01) \\
   & Power & 0.43(0.05) & 0.54(0.04) & 0.66(0.01) & 0.67(0.02) & 0.47(0.01) & 0.48(0.01) & 0.41(0.04) & 0.62(0.02) & 0.40(0.03) \\
\hline
ML & FDR   & 0.02(0.01) & 0.07(0.04) & 0.07(0.03) & 0.06(0.03) & 0.08(0.03) & 0.05(0.05) & 0.07(0.01) & 0.03(0.04) & 0.10(0.02) \\
   & Power & 0.61(0.03) & 0.46(0.03) & 0.45(0.00) & 0.52(0.04) & 0.51(0.01) & 0.50(0.05) & 0.57(0.01) & 0.54(0.04) & 0.66(0.04) \\
\hline
VWA-OL & FDR   & 0.02(0.03) & 0.02(0.01) & 0.04(0.04) & 0.04(0.05) & 0.03(0.00) & 0.04(0.05) & 0.03(0.04) & 0.02(0.01) & 0.05(0.04) \\
        & Power & 0.59(0.02) & 0.53(0.05) & 0.63(0.03) & 0.47(0.05) & 0.61(0.02) & 0.51(0.05) & 0.48(0.02) & 0.70(0.03) & 0.62(0.04) \\
\hline
VWA-ML & FDR   & 0.01(0.04) & 0.03(0.02) & 0.04(0.03) & 0.02(0.02) & 0.04(0.03) & 0.03(0.00) & 0.03(0.04) & 0.03(0.01) & 0.02(0.00) \\
        & Power & 0.63(0.03) & 0.49(0.02) & 0.62(0.00) & 0.48(0.01) & 0.49(0.02) & 0.50(0.04) & 0.68(0.04) & 0.41(0.02) & 0.52(0.01) \\
\hline
VWA-OML & FDR   & 0.04(0.01) & 0.03(0.04) & 0.04(0.03) & 0.01(0.02) & 0.02(0.01) & 0.01(0.00) & 0.01(0.00) & 0.02(0.03) & 0.04(0.02) \\
         & Power & 0.57(0.04) & 0.51(0.00) & 0.55(0.03) & 0.43(0.01) & 0.49(0.01) & 0.53(0.02) & 0.66(0.01) & 0.64(0.01) & 0.64(0.03) \\
\hline
\end{tabular}
}
\vspace{0.6em}
\caption{Filtered results for normal binary classification under normal binary distribution with varied network structures and related variable ratios, target FDR q = 0.1.}
\label{tab:normal_binary}
\end{table}

\begin{table}[h]
\scriptsize
\centering
\setlength{\tabcolsep}{6pt}
\renewcommand{\arraystretch}{1.1}
\resizebox{\textwidth}{!}{  
\begin{tabular}{|c|c||*{9}{c|}}
\hline
\multicolumn{2}{|c||}{Model structure} & \multicolumn{3}{c|}{(50,64,1)} & \multicolumn{3}{c|}{(100,264,1)} & \multicolumn{3}{c|}{(300,264,1)} \\
\hline
\multicolumn{2}{|c||}{Distribution} & \multicolumn{9}{c|}{Normal distribution, mean = 0, std = 1} \\
\hline
\multicolumn{2}{|c||}{Total variables} & \multicolumn{3}{c|}{25 variables} & \multicolumn{3}{c|}{50 variables} & \multicolumn{3}{c|}{150 variables} \\
\hline
\multicolumn{2}{|c||}{Related variables} & 5 & 10 & 20 & 10 & 25 & 40 & 100 & 150 & 200 \\
\hline
OL & FDR   & 0.08(0.01) & 0.04(0.03) & 0.09(0.04) & 0.05(0.01) & 0.07(0.00) & 0.08(0.02) & 0.07(0.03) & 0.03(0.01) & 0.02(0.01) \\
   & Power & 0.59(0.01) & 0.43(0.01) & 0.64(0.03) & 0.60(0.01) & 0.48(0.01) & 0.44(0.02) & 0.58(0.05) & 0.44(0.00) & 0.43(0.01) \\
\hline
ML & FDR   & 0.08(0.01) & 0.05(0.03) & 0.05(0.02) & 0.08(0.03) & 0.03(0.05) & 0.02(0.00) & 0.10(0.04) & 0.09(0.04) & 0.05(0.01) \\
   & Power & 0.52(0.00) & 0.49(0.04) & 0.64(0.00) & 0.55(0.03) & 0.54(0.00) & 0.55(0.02) & 0.52(0.01) & 0.56(0.05) & 0.63(0.02) \\
\hline
VWA-OL & FDR   & 0.04(0.02) & 0.02(0.00) & 0.04(0.03) & 0.04(0.00) & 0.02(0.01) & 0.02(0.04) & 0.03(0.00) & 0.03(0.00) & 0.02(0.02) \\
        & Power & 0.55(0.04) & 0.43(0.04) & 0.46(0.05) & 0.48(0.01) & 0.45(0.01) & 0.64(0.05) & 0.63(0.04) & 0.58(0.03) & 0.63(0.03) \\
\hline
VWA-ML & FDR   & 0.02(0.03) & 0.02(0.05) & 0.03(0.04) & 0.02(0.00) & 0.02(0.02) & 0.02(0.03) & 0.02(0.00) & 0.04(0.02) & 0.03(0.03) \\
        & Power & 0.68(0.05) & 0.72(0.02) & 0.71(0.04) & 0.60(0.04) & 0.64(0.01) & 0.58(0.01) & 0.71(0.00) & 0.67(0.05) & 0.59(0.03) \\
\hline
VWA-OML & FDR   & 0.02(0.05) & 0.03(0.00) & 0.01(0.00) & 0.04(0.03) & 0.04(0.05) & 0.01(0.01) & 0.03(0.02) & 0.03(0.03) & 0.04(0.01) \\
         & Power & 0.70(0.03) & 0.43(0.01) & 0.71(0.04) & 0.55(0.01) & 0.74(0.01) & 0.57(0.04) & 0.59(0.04) & 0.72(0.02) & 0.63(0.04) \\
\hline
\end{tabular}
}
\vspace{0.6em}
\caption{Filtered results for normal muti classification under normal muti distribution with varied network structures and related variable ratios, target FDR q = 0.1.}
\label{tab:normal_muti}
\end{table}

\begin{table}[h]
\scriptsize
\centering
\setlength{\tabcolsep}{6pt}
\renewcommand{\arraystretch}{1.1}
\resizebox{\textwidth}{!}{  
\begin{tabular}{|c|c||*{9}{c|}} 
\hline
\multicolumn{2}{|c||}{Model structure} 
& \multicolumn{3}{c|}{(50,64,1)} 
& \multicolumn{3}{c|}{(100,264,1)} 
& \multicolumn{3}{c|}{(300,264,1)} \\
\hline
\multicolumn{2}{|c||}{Distribution} 
& \multicolumn{9}{c|}{Chi-square distribution, df = 4} \\
\hline
\multicolumn{2}{|c||}{Total variables} 
& \multicolumn{3}{c|}{25 variables} 
& \multicolumn{3}{c|}{50 variables} 
& \multicolumn{3}{c|}{150 variables} \\
\hline
\multicolumn{2}{|c||}{Related variables} 
& 5 & 10 & 20 & 10 & 25 & 40 & 100 & 150 & 200 \\
\hline
OL & FDR   & 0.17(0.02) & 0.22(0.04) & 0.19(0.01) & 0.07(0.02) & 0.07(0.03) & 0.11(0.00) & 0.07(0.04) & 0.10(0.05) & 0.20(0.04) \\
   & Power & 0.50(0.05) & 0.44(0.04) & 0.49(0.04) & 0.44(0.02) & 0.45(0.02) & 0.50(0.03) & 0.43(0.02) & 0.59(0.05) & 0.51(0.05) \\
\hline
ML & FDR   & 0.17(0.02) & 0.21(0.01) & 0.12(0.01) & 0.07(0.03) & 0.09(0.03) & 0.08(0.03) & 0.06(0.02) & 0.10(0.01) & 0.20(0.03) \\
   & Power & 0.56(0.02) & 0.58(0.04) & 0.56(0.01) & 0.42(0.03) & 0.53(0.02) & 0.60(0.02) & 0.52(0.03) & 0.45(0.03) & 0.47(0.05) \\
\hline
VWA-OL & FDR   & 0.14(0.03) & 0.14(0.03) & 0.09(0.03) & 0.11(0.04) & 0.11(0.01) & 0.06(0.02) & 0.15(0.01) & 0.13(0.03) & 0.11(0.01) \\
        & Power & 0.61(0.04) & 0.55(0.05) & 0.58(0.02) & 0.63(0.03) & 0.58(0.04) & 0.57(0.00) & 0.59(0.00) & 0.56(0.03) & 0.58(0.04) \\
\hline
VWA-ML & FDR   & 0.12(0.02) & 0.08(0.03) & 0.10(0.03) & 0.14(0.02) & 0.14(0.04) & 0.11(0.01) & 0.06(0.01) & 0.08(0.01) & 0.12(0.05) \\
        & Power & 0.58(0.01) & 0.58(0.03) & 0.56(0.00) & 0.62(0.01) & 0.61(0.00) & 0.63(0.01) & 0.64(0.02) & 0.58(0.02) & 0.63(0.04) \\
\hline
VWA-OML & FDR   & 0.14(0.02) & 0.07(0.02) & 0.14(0.02) & 0.10(0.03) & 0.15(0.01) & 0.11(0.04) & 0.13(0.02) & 0.13(0.00) & 0.13(0.01) \\
         & Power & 0.65(0.01) & 0.58(0.02) & 0.57(0.04) & 0.64(0.01) & 0.60(0.04) & 0.63(0.01) & 0.62(0.05) & 0.60(0.03) & 0.64(0.04) \\
\hline
\end{tabular}
}
\vspace{0.6em}
\caption{Filtered results for binary classification under chi-square distribution with varied network structures and related variable ratios, target FDR $q = 0.1$.}
\label{tab12a}
\end{table}

\begin{table}[h]
\scriptsize
\centering
\setlength{\tabcolsep}{6pt}
\renewcommand{\arraystretch}{1.1}
\resizebox{\textwidth}{!}{  
\begin{tabular}{|c|c||*{9}{c|}} 
\hline
\multicolumn{2}{|c||}{Model structure} 
& \multicolumn{3}{c|}{(50,64,3)} 
& \multicolumn{3}{c|}{(100,264,3)} 
& \multicolumn{3}{c|}{(300,264,3)} \\
\hline
\multicolumn{2}{|c||}{Distribution} 
& \multicolumn{9}{c|}{Chi-square distribution, df = 4} \\
\hline
\multicolumn{2}{|c||}{Total variables} 
& \multicolumn{3}{c|}{25 variables} 
& \multicolumn{3}{c|}{50 variables} 
& \multicolumn{3}{c|}{150 variables} \\
\hline
\multicolumn{2}{|c||}{Related variables} 
& 5 & 10 & 20 & 10 & 25 & 40 & 100 & 150 & 200 \\
\hline
OL & FDR   & 0.09(0.04) & 0.18(0.02) & 0.19(0.00) & 0.18(0.02) & 0.16(0.02) & 0.18(0.01) & 0.17(0.04) & 0.16(0.03) & 0.09(0.04) \\
   & Power & 0.47(0.01) & 0.52(0.03) & 0.51(0.02) & 0.47(0.04) & 0.49(0.05) & 0.51(0.02) & 0.53(0.03) & 0.45(0.00) & 0.47(0.02) \\
\hline
ML & FDR   & 0.08(0.03) & 0.18(0.04) & 0.19(0.04) & 0.16(0.04) & 0.18(0.04) & 0.10(0.02) & 0.07(0.03) & 0.07(0.01) & 0.13(0.03) \\
   & Power & 0.49(0.01) & 0.41(0.04) & 0.47(0.02) & 0.44(0.01) & 0.46(0.04) & 0.51(0.01) & 0.54(0.04) & 0.52(0.03) & 0.40(0.00) \\
\hline
VWA-OL & FDR   & 0.09(0.05) & 0.09(0.00) & 0.09(0.04) & 0.08(0.04) & 0.07(0.01) & 0.10(0.03) & 0.06(0.02) & 0.10(0.04) & 0.09(0.03) \\
        & Power & 0.47(0.04) & 0.57(0.02) & 0.55(0.03) & 0.66(0.01) & 0.48(0.02) & 0.56(0.05) & 0.43(0.03) & 0.52(0.03) & 0.53(0.01) \\
\hline
VWA-ML & FDR   & 0.07(0.05) & 0.07(0.00) & 0.08(0.03) & 0.06(0.04) & 0.07(0.01) & 0.09(0.01) & 0.09(0.01) & 0.10(0.04) & 0.08(0.01) \\
        & Power & 0.45(0.02) & 0.65(0.01) & 0.42(0.03) & 0.54(0.02) & 0.52(0.02) & 0.52(0.05) & 0.54(0.04) & 0.42(0.01) & 0.52(0.04) \\
\hline
VWA-OML & FDR   & 0.08(0.01) & 0.06(0.03) & 0.08(0.01) & 0.10(0.05) & 0.09(0.04) & 0.08(0.03) & 0.06(0.00) & 0.08(0.05) & 0.08(0.00) \\
         & Power & 0.45(0.04) & 0.45(0.04) & 0.57(0.01) & 0.51(0.03) & 0.60(0.03) & 0.56(0.01) & 0.58(0.02) & 0.62(0.01) & 0.60(0.04) \\
\hline
\end{tabular}
}
\vspace{0.6em}
\caption{Filtered results for multi-classification under chi-square distribution with varied network structures and related variable ratios, target FDR $q = 0.1$.}
\label{tab12b}
\end{table}

\begin{table}[h]
\scriptsize
\centering
\setlength{\tabcolsep}{6pt}
\renewcommand{\arraystretch}{1.1}
\resizebox{\textwidth}{!}{  
\begin{tabular}{|c|c||*{9}{c|}}
\hline
\multicolumn{2}{|c||}{Model structure} & \multicolumn{3}{c|}{(50,64,1)} & \multicolumn{3}{c|}{(100,264,1)} & \multicolumn{3}{c|}{(300,264,1)} \\
\hline
\multicolumn{2}{|c||}{Distribution} & \multicolumn{9}{c|}{Gamma distribution, $\alpha=2, \beta=1$} \\
\hline
\multicolumn{2}{|c||}{Total variables} & \multicolumn{3}{c|}{25 variables} & \multicolumn{3}{c|}{50 variables} & \multicolumn{3}{c|}{150 variables} \\
\hline
\multicolumn{2}{|c||}{Related variables} & 5 & 10 & 20 & 10 & 25 & 40 & 100 & 150 & 200 \\
\hline
OL & FDR   & 0.17(0.03) & 0.17(0.04) & 0.06(0.04) & 0.06(0.00) & 0.16(0.05) & 0.16(0.05) & 0.09(0.02) & 0.09(0.04) & 0.14(0.04) \\
   & Power & 0.35(0.03) & 0.35(0.02) & 0.47(0.03) & 0.46(0.03) & 0.49(0.01) & 0.31(0.04) & 0.32(0.02) & 0.48(0.03) & 0.43(0.01) \\
\hline
ML & FDR   & 0.08(0.02) & 0.05(0.04) & 0.23(0.05) & 0.11(0.02) & 0.09(0.04) & 0.12(0.05) & 0.21(0.01) & 0.08(0.01) & 0.09(0.04) \\
   & Power & 0.38(0.03) & 0.38(0.04) & 0.36(0.03) & 0.46(0.04) & 0.44(0.02) & 0.50(0.04) & 0.43(0.00) & 0.50(0.01) & 0.45(0.02) \\
\hline
VWA-OL & FDR   & 0.19(0.03) & 0.18(0.01) & 0.11(0.03) & 0.12(0.05) & 0.20(0.03) & 0.20(0.02) & 0.10(0.05) & 0.14(0.01) & 0.11(0.02) \\
        & Power & 0.51(0.03) & 0.45(0.03) & 0.40(0.01) & 0.48(0.03) & 0.47(0.02) & 0.46(0.04) & 0.43(0.01) & 0.55(0.04) & 0.50(0.04) \\
\hline
VWA-ML & FDR   & 0.14(0.05) & 0.12(0.02) & 0.16(0.05) & 0.11(0.02) & 0.14(0.00) & 0.12(0.04) & 0.12(0.01) & 0.17(0.01) & 0.11(0.00) \\
        & Power & 0.58(0.02) & 0.53(0.03) & 0.45(0.01) & 0.48(0.05) & 0.49(0.03) & 0.48(0.03) & 0.50(0.02) & 0.51(0.02) & 0.55(0.01) \\
\hline
VWA-OML & FDR   & 0.13(0.04) & 0.12(0.03) & 0.20(0.02) & 0.10(0.01) & 0.17(0.04) & 0.19(0.01) & 0.13(0.01) & 0.17(0.04) & 0.14(0.00) \\
         & Power & 0.57(0.03) & 0.41(0.04) & 0.47(0.04) & 0.59(0.01) & 0.54(0.01) & 0.42(0.01) & 0.46(0.05) & 0.41(0.00) & 0.60(0.04) \\
\hline
\end{tabular}
}
\vspace{0.6em}
\caption{Filtered results for gamma binary classification under gamma binary distribution with varied network structures and related variable ratios, target FDR q = 0.1.}
\label{tab:gamma_binary}
\end{table}

\begin{table}[h]
\scriptsize
\centering
\setlength{\tabcolsep}{6pt}
\renewcommand{\arraystretch}{1.1}
\resizebox{\textwidth}{!}{  
\begin{tabular}{|c|c||*{9}{c|}}
\hline
\multicolumn{2}{|c||}{Model structure} & \multicolumn{3}{c|}{(50,64,3)} & \multicolumn{3}{c|}{(100,264,3)} & \multicolumn{3}{c|}{(300,264,3)} \\
\hline
\multicolumn{2}{|c||}{Distribution} & \multicolumn{9}{c|}{Gamma distribution, $\alpha=2, \beta=1$} \\
\hline
\multicolumn{2}{|c||}{Total variables} & \multicolumn{3}{c|}{25 variables} & \multicolumn{3}{c|}{50 variables} & \multicolumn{3}{c|}{150 variables} \\
\hline
\multicolumn{2}{|c||}{Related variables} & 5 & 10 & 20 & 10 & 25 & 40 & 100 & 150 & 200 \\
\hline
OL & FDR   & 0.20(0.02) & 0.28(0.03) & 0.10(0.04) & 0.07(0.02) & 0.25(0.01) & 0.14(0.01) & 0.17(0.04) & 0.25(0.03) & 0.30(0.00) \\
   & Power & 0.54(0.02) & 0.47(0.04) & 0.44(0.03) & 0.48(0.04) & 0.45(0.05) & 0.51(0.03) & 0.42(0.01) & 0.47(0.03) & 0.48(0.04) \\
\hline
ML & FDR   & 0.17(0.01) & 0.06(0.01) & 0.06(0.03) & 0.29(0.02) & 0.26(0.01) & 0.16(0.03) & 0.20(0.01) & 0.10(0.01) & 0.23(0.05) \\
   & Power & 0.49(0.00) & 0.43(0.04) & 0.45(0.05) & 0.50(0.01) & 0.45(0.04) & 0.52(0.02) & 0.53(0.01) & 0.43(0.01) & 0.51(0.03) \\
\hline
VWA-OL & FDR   & 0.21(0.01) & 0.14(0.04) & 0.21(0.05) & 0.15(0.02) & 0.13(0.04) & 0.11(0.00) & 0.12(0.05) & 0.23(0.01) & 0.23(0.01) \\
        & Power & 0.64(0.04) & 0.44(0.01) & 0.52(0.05) & 0.59(0.01) & 0.49(0.00) & 0.45(0.04) & 0.58(0.02) & 0.55(0.01) & 0.42(0.04) \\
\hline
VWA-ML & FDR   & 0.11(0.04) & 0.11(0.00) & 0.24(0.03) & 0.19(0.00) & 0.18(0.02) & 0.25(0.04) & 0.13(0.02) & 0.16(0.01) & 0.16(0.01) \\
        & Power & 0.41(0.01) & 0.44(0.03) & 0.60(0.04) & 0.55(0.03) & 0.52(0.00) & 0.49(0.04) & 0.61(0.03) & 0.51(0.04) & 0.61(0.02) \\
\hline
VWA-OML & FDR   & 0.24(0.03) & 0.23(0.01) & 0.25(0.02) & 0.10(0.04) & 0.25(0.02) & 0.13(0.02) & 0.13(0.03) & 0.24(0.02) & 0.13(0.00) \\
         & Power & 0.51(0.02) & 0.47(0.05) & 0.42(0.02) & 0.64(0.02) & 0.43(0.03) & 0.54(0.02) & 0.62(0.02) & 0.65(0.04) & 0.40(0.04) \\
\hline
\end{tabular}
}
\vspace{0.6em}
\caption{Filtered results for gamma muti classification under gamma muti distribution with varied network structures and related variable ratios, target FDR q = 0.1.}
\label{tab:gamma_muti}
\end{table}

\end{document}